\documentclass[journal]{IEEEtran}
\IEEEoverridecommandlockouts

\usepackage[hidelinks]{hyperref}
\usepackage[cmex10]{amsmath}
\usepackage{amssymb,amsfonts}
\usepackage{dblfloatfix}

\usepackage[ruled,vlined]{algorithm2e}
\usepackage{graphicx}
\graphicspath{{Figures/PDF/}{Figures/PNG/}}

\usepackage{booktabs}
\usepackage{siunitx}
\usepackage[numbers,compress]{natbib}
\usepackage{texnames}
\usepackage{soul}
\usepackage{array}
\usepackage{bm,bbm}
\usepackage{orcidlink}

\usepackage{csquotes}
\usepackage{multirow}
\usepackage{comment}
\usepackage{longtable}
\usepackage{ragged2e}
\newcolumntype{P}[1]{>{\RaggedRight\arraybackslash}p{#1}}

\usepackage{subcaption}
\usepackage{tabularx}

\begin{document}
\title{SULAND\_v2: A Refined RGB Dataset and Deep Learning Object Detection Benchmark for UAV/UGV-Based SUrface LANDmine Detection Under Domain Shift}

\author{Sagar Lekhak$^{1\dagger}$~\orcidlink{0009-0009-7896-6167},
Prasanna Reddy Pulakurthi$^{1\dagger}$~\orcidlink{0000-0003-0486-0756},
Lalit Joshi$^{2}$, Ramesh Bhatta$^{1}$, and
Emmett J. Ientilucci$^{1}$~\orcidlink{0000-0002-3643-8245}%
\thanks{$^{1}$Sagar Lekhak, Prasanna Reddy Pulakurthi, Ramesh Bhatta, and Emmett J. Ientilucci are with the Rochester Institute of Technology, Rochester, NY 14623, USA.}%
\thanks{$^{2}$Lalit Joshi is with Thapathali Campus, Institute of Engineering, Tribhuvan University, Kathmandu 44600, Nepal.}%
\thanks{Corresponding author: Sagar Lekhak (email: sl3088@rit.edu).}%
\thanks{$^{\dagger}$These authors contributed equally to this work.}}

\maketitle
\begin{abstract}

RGB imagery offers a practical, low-cost option for Unmanned Aerial/Ground Vehicle (UAV/UGV) survey support in surface-landmine detection, but object detectors remain underexplored in this safety-critical domain. Limited cross-architecture benchmarking and insufficient out-of-distribution (OOD) analysis make it difficult to assess whether detectors generalize across deployment conditions. This challenge is amplified by the scarcity of public RGB landmine datasets for domain-shift evaluation, making SULAND dataset an important benchmark for PFM-1 and PMA-2 detection. However, our inspection of SULAND revealed missing and false annotations, localization errors, inconsistent visibility criteria, visual artifacts, temporal labeling inconsistencies, and an inverted OOD class-ID convention. This paper presents SULAND\_v2, a refined RGB surface-landmine dataset and object-detection benchmark. The original images and splits are preserved, while annotations are manually revised to improve completeness, localization, label validity, and class consistency. SULAND\_v2 contains 33{,}771 images and 12{,}433 bounding-box annotations. We quantify the effect of refinement through cross-version evaluation and benchmark 35 detector configurations across nine detector families. Annotation refinement improves YOLOv8 in-distribution (IID) test mAP@50 by 14.6--19.6 percentage points, while correcting the OOD class-ID convention increases mean YOLOv8 OOD mAP@50 by approximately 25 percentage points. On SULAND\_v2, YOLOv12-Small achieves the highest IID mAP@50 (0.908), whereas RF-DETR-Large achieves the strongest OOD performance (0.799 mAP@50, 0.675 recall). These results show that high IID accuracy is not sufficient evidence of operational readiness.
With its corrected annotations, preserved IID/OOD splits, and extensive baseline evaluations across diverse detector families, the refined SULAND\_v2 represents a highly consistent benchmark for studying domain-shift robustness in RGB-based mine-action survey support.

\end{abstract}

\begin{IEEEkeywords}
Surface landmine detection, UAV remote sensing, RGB imagery, object detection, deep learning, dataset refinement, benchmark dataset, out-of-distribution generalization, domain shift, YOLO, SULAND.
\end{IEEEkeywords}

\section{Introduction}
\label{sec:introduction}

\IEEEPARstart{L}{andmines} and other explosive ordnance remain a persistent humanitarian and developmental challenge in post-conflict and active-conflict regions. Beyond direct casualties, explosive contamination restricts access to agricultural land, transportation corridors, schools, water sources, and critical infrastructure, while also slowing reconstruction and the safe return of displaced communities \cite{unmas_mine_action, gichd_detection_clearance}. According to \emph{Landmine Monitor 2025}, at least 6,279 people were killed or injured by landmines and explosive remnants of war in 2024, with civilians accounting for 90\% of recorded casualties where status was known \cite{landmine_monitor_2025}. In conflict-affected areas, including regions where civilian infrastructure has been damaged during the Ukraine--Russia war, explosive contamination also obstructs agricultural recovery and safe land release, motivating faster survey and decision-support technologies \cite{halo_ukraine, gichd_survey}.

Humanitarian demining is essential, but it remains slow, hazardous, and resource-intensive because clearance requires high-confidence inspection of suspected hazardous areas, often under difficult terrain, vegetation, access, and safety constraints \cite{gichd_detection_clearance, gichd_survey}. Operational Mine Action therefore relies on evidence-based land release, beginning with non-technical survey (NTS), followed where required by technical survey (TS) and targeted clearance \cite{gichd_survey, gichd_nts_guide}. Remote sensing and computer vision can enhance these workflows by providing decision-support information for prioritizing suspected hazardous areas, improving situational awareness, and reducing the area and time required for high-risk or high-cost inspection.

Unmanned aerial vehicles (UAVs) and Unmanned Ground Vehicles (UGVs) are attractive for this role because they provide flexible, low-altitude, high-spatial-resolution data collection over areas that may be unsafe or inefficient to survey manually \cite{2014_COLOMINA_uav_review, 2014_nex}. UAV-based remote sensing for mine action has explored thermal, multispectral, hyperspectral, RGB, radar, magnetometer, electromagnetic, and metal-detection approaches \cite{2018_nikulin_thermal, 2017_Cerquera_UAV_GPR, Barnawi2022AChallenges, Lekhak2024ViabilityDetection, baur2026comparative, lekhak2026humanintheloopsignaturebootstrappinguav}. Among these modalities, RGB imagery is particularly practical for UAV-based mine-action workflows because cameras are inexpensive, lightweight, widely available, and readily compatible with modern object-detection pipelines.

Recent studies have explored deep-learning-based object detection in UAV-based RGB imagery, including Faster R-CNN, YOLO-family models, and multimodal detection approaches for landmine and unexploded ordnance detection \cite{baur2021how_to_implement_drones, Vivoli2024DeepImaging, joint_fusion_and_detection_2023}. These works show that deep detectors can improve survey speed and detection accuracy when training and test data share similar distributions, such as comparable backgrounds, viewpoints, and acquisition conditions. This represents an in-distribution (IID) setting. However, under domain shift, detectors may encounter out-of-distribution (OOD) settings in which the test distribution differs from the training distribution due to changes in viewpoint, background, target appearance, or acquisition conditions. Similar concerns have been reported in broader computer-vision and remote-sensing object-detection applications, where models can degrade substantially under real-world distribution shifts \cite{lyu2025deep, jstars_da, tgrs_da, alemadi2025rwds}.

OOD evaluation is especially important for surface landmine detection because real survey environments are rarely controlled. Surface mines may appear across roads, gravel, grass, agricultural fields, different soil types, and varying vegetation states, while their appearance can change with illumination, camera geometry, target distance, geographic region, and partial occlusion. Visible surface targets are also often small, visually subtle, and easily confused with natural clutter such as shadows, stones, soil texture, vegetation, and background objects. A detector trained in one environment may therefore fail to generalize to another, even when the target class is nominally unchanged. In safety-critical tasks such as landmine detection, IID performance alone is therefore insufficient; detector behavior under OOD conditions must be evaluated explicitly.

Despite rapid progress in object detection, many detector families remain underexplored for RGB-based surface landmine detection under a common IID/OOD evaluation protocol. Conclusions drawn from a narrow set of models may not adequately characterize the strengths and limitations of current detectors for small, rare, visually subtle, and safety-critical targets. This creates an important benchmark question for the demining research community: \textit{whether detectors that perform well on familiar data can maintain reliable target recovery under new environmental conditions relevant to technical and non-technical survey workflows.}

This question is difficult to answer because publicly available UAV-based or low-altitude RGB datasets for surface landmine detection remain scarce. Unlike general object detection, where large-scale benchmarks support repeated comparison across methods, landmine detection research relies on a small number of public datasets, many of which are limited in geography, target type, environmental diversity, annotation detail, or evaluation protocol. Collecting imagery with realistic mine targets is constrained by safety, access, and security considerations; therefore, datasets based on inert mines, surrogates, or 3D-printed replicas are valuable community resources. Their scarcity also makes each released benchmark disproportionately influential, since a single dataset can shape which methods are developed, compared, and trusted.

Dataset quality is therefore central to credible progress. Annotation errors, including missing labels, false positives, inaccurate bounding boxes, inconsistent partial-object criteria, and class-map mismatches, can significantly distort both training behavior and reported evaluation results \cite{schubert2023labelerrors}. These issues are particularly consequential under OOD evaluation because apparent generalization failures may reflect benchmark noise rather than genuine detector limitations. For scarce-data, safety-critical remote-sensing tasks, benchmark reliability is therefore not a secondary concern but a prerequisite for meaningful detector comparison.

Among available RGB resources, the SULAND dataset \cite{Vivoli2024DeepImaging}, hereafter referred to as SULAND\_v1, is particularly important because it provides low-altitude RGB imagery of realistic mine surrogates and separates IID and OOD evaluation settings. The dataset includes visually challenging PFM-1 and PMA-2 targets collected under varied environmental conditions. The IID data include targets over grass and gravel surfaces under sunny, cloudy, and shadowed conditions, with scene elements such as bushes, branches, walls, bars, tree trunks, and rocks. The OOD sequences introduce a geographic domain shift from Italy to the United States and includes additional variation in environmental context, slope, camera viewpoint, target distance, partial occlusion, target size, color, and appearance. This structure makes SULAND\_v1 a valuable baseline for studying detector robustness in RGB-based surface landmine detection.

However, our inspection of the released SULAND\_v1 annotations revealed multiple data-quality and benchmark-consistency issues, including missing labels, false annotations, inconsistent temporal labeling, mislocalized boxes, ambiguous annotation criteria, and a class-ID convention mismatch in the OOD labels. These issues can affect both model training and evaluation, making it difficult to distinguish genuine detector limitations from dataset-induced artifacts. A systematic reassessment of SULAND is therefore necessary before it can serve as a reliable benchmark for IID and OOD evaluation.

In this paper, we revisit SULAND\_v1 from the perspective of benchmark reliability. We manually audit the original dataset, categorize the observed annotation and benchmark issues, construct a refined and consistently annotated version named SULAND\_v2, and evaluate a broad set of object detectors under both IID and OOD settings. In contrast to using SULAND\_v1 only as a source dataset for detector evaluation, this work treats dataset refinement, benchmark validation, and detector comparison as a unified problem. The goal is not only to improve a scarce public dataset, but also to clarify how annotation quality and evaluation consistency influence conclusions about detector performance in RGB-based surface landmine detection.

The main contributions of this paper are summarized as follows:

\begin{itemize}

\item \textbf{Systematic audit of SULAND\_v1:}
We conduct a folder-by-folder and frame-by-frame audit of the released SULAND\_v1 images and annotations. The audit identifies missing and invalid annotations, mislocalized bounding boxes, inconsistent partial-visibility criteria, non-representative artifacts, temporal annotation inconsistencies, and an inverted OOD class-ID convention.

\item \textbf{Construction of SULAND\_v2:}
We refine the complete dataset using unified annotation criteria, consistent class definitions, corrected bounding boxes, and harmonized IID/OOD class conventions, while preserving the original imagery and split organization. The resulting SULAND\_v2 release provides a more consistent basis for RGB-based surface-mine detection research.

\item \textbf{Quantitative assessment of annotation refinement:}
We characterize the changes from SULAND\_v1 to SULAND\_v2 using split-level statistics, annotation-revision measurements, representative correction examples, and cross-version training and evaluation. We additionally isolate the effect of the inverted OOD class convention by re-evaluating the same predictions using corrected class IDs.

\item \textbf{Broad object-detector benchmark:}
We evaluate 35 model configurations from nine detector families on SULAND\_v2, including one-stage, two-stage, transformer-based, and vision--language-based object detectors. All models are assessed using a common evaluation protocol that reports detection accuracy, precision, recall, parameter count, and inference speed.

\item \textbf{IID--OOD robustness and efficiency analysis:}
We examine changes in detector ranking between IID and OOD conditions, class-specific precision and recall, and the tradeoff among detection accuracy, inference speed, and OOD robustness. This analysis identifies limitations that are not apparent from IID performance alone and provides guidance for selecting detectors under different computational and evaluation requirements.

\end{itemize}

To the best of our knowledge, this is the first study to systematically audit, categorize, and refine the SULAND dataset while jointly evaluating benchmark reliability, detector performance, and OOD robustness for RGB-based surface landmine detection.

The remainder of this paper is organized as follows. Section~\ref{sec:literature_review} reviews RGB-based surface-mine detection, relevant UAV/UGV datasets, and existing gaps in benchmark reliability and domain-shift evaluation. Section~\ref{sec:suland_v1} describes SULAND\_v1, its IID/OOD organization, and the annotation-audit procedure and findings. Section~\ref{sec:suland_v2} presents the construction of SULAND\_v2, the quantitative and qualitative annotation changes, and the same- and cross-version evaluation. Section~\ref{sec:benchmark_design_experimental_setup} describes the evaluated detector families, experimental protocol, and evaluation metrics. Section~\ref{sec:detector_results} reports the overall benchmark results, IID--OOD generalization gap, accuracy--speed tradeoff, class-wise performance, and qualitative error analysis. Section~\ref{sec:discussion} discusses benchmark reliability, robustness evaluation, operational relevance, limitations, and future directions. Finally, Section~\ref{sec:conclusion} summarizes the principal findings, and Section~\ref{sec:dataset_code_availability} provides dataset and code availability information.

\section{Literature Review}
\label{sec:literature_review}

\subsection{UAV/UGV-Based Remote Sensing and RGB Imagery for Mine Action}

UAVs and UGVs have become important platforms for high-resolution sensing because they enable flexible, low-altitude, and repeatable data acquisition over areas that may be difficult or unsafe to access directly \cite{2014_COLOMINA_uav_review, 2014_nex}. These properties are relevant to mine action, where remote sensing is generally intended to support survey, prioritization, and risk reduction rather than replace manual or mechanical clearance. Compared with satellite or conventional airborne platforms, UAVs can collect centimeter-scale imagery over suspected hazardous areas, while UGVs can acquire closer-range observations with more controlled sensor-target geometry.

UAV- and UGV-based mine-action sensing has been investigated using several modalities, including RGB, thermal infrared, multispectral, hyperspectral, LiDAR, synthetic aperture radar (SAR), electromagnetic induction (EMI), magnetometry, and ground-penetrating radar (GPR) \cite{2018_nikulin_thermal, 2017_Cerquera_UAV_GPR, Barnawi2022AChallenges, Lekhak2024ViabilityDetection, baur2026comparative}. These modalities address different detection requirements. Thermal and spectral sensors can exploit temperature, material, soil, or vegetation-related contrast, whereas geophysical methods such as EMI, magnetometry, and GPR are more directly relevant to metallic or buried targets \cite{2018_nikulin_thermal, 2017_Cerquera_UAV_GPR,  Makki2017AImaging, Lekhak2026ADetection, Lekhak2026BenchmarkingImagery, Kuru2025AutomatedRadar}. In contrast, RGB-based detection is primarily applicable to visible surface targets, where mine-like objects are localized from visual appearance, spatial context, and background contrast.

Comparative multimodal studies further show that sensor effectiveness depends strongly on target type, target visibility, material composition, and deployment configuration. For example, Baur \textit{et al}.~\cite{baur2026comparative} evaluated multiple UAV- and ground-based sensing modalities over a standardized seeded minefield and reported that RGB imagery achieved the highest detection rate for visible surface objects, whereas geophysical methods were more relevant for metallic or buried targets. This finding supports the use of RGB imagery for surface mine detection, particularly because it provides high-resolution visual evidence that can assist both automated object detection and human interpretation. RGB-based detection is therefore especially relevant for rapid survey support in open or semi-open environments where surface ordnance is visible or partially visible. Accordingly, the present study focuses on RGB-based surface mine detection and the benchmark protocols used to evaluate detector robustness under domain shift.

\subsection{Deep Learning in UAV/UGV-Based RGB Imagery for Surface Landmine Detection }
\label{subsec:deep_learning_sruf_landmine_det}

The use of RGB imagery for surface landmine detection is closely related to progress in deep learning-based object detection. Two-stage detectors such as Faster R-CNN \cite{2015_FasterRCNN} and one-stage detectors such as YOLO \cite{2016_yolo} established practical frameworks for localizing objects in images. These detector families have also been widely adopted in aerial and UAV imagery, where small objects, nonuniform scale, oblique viewpoints, and cluttered backgrounds are common challenges \cite{2018_dota_dataset, 2018uavdt_dataset, 2018vis-drone_dataset, 2018_DENG}. These same challenges are central to RGB-based surface mine detection, where targets may occupy a small image region and may visually resemble soil, stones, shadows, vegetation, or other background structures.

Several studies have applied deep learning-based classification and object-detection models to RGB imagery for mine and UXO detection. Baur \textit{et al.}~\cite{baur2021how_to_implement_drones, rs12050859} applied Faster R-CNN to UAV-acquired imagery for PFM-1 detection and reported strong performance under partially withheld testing, with accuracy decreasing from 99.3\% on a partially withheld test set to 71.5\% on a completely withheld test set. Their study also showed that detection performance degraded when mines were partially buried or when mine surfaces were occluded by vegetation. Qiu \textit{et al}.~\cite{joint_fusion_and_detection_2023} proposed an RGB--near-infrared YOLOv5-based fusion framework for detecting several mine types, but reported false positives across multiple experimental scenes. Agrawal-Chung and Moin~\cite{agrawalchung2024dronflyby} evaluated YOLOF, DETR~\cite{carion2020end}, Sparse R-CNN~\cite{sun2021sparse}, and VarifocalNet~\cite{zhang2021varifocalnet} for drone-based surface mine detection across three deployment altitudes, demonstrating the feasibility of modern object detectors for accelerating the detection process. Lekhak \textit{et al}.~\cite{Lekhak2025UncertaintyDropout} introduced a Monte Carlo dropout-based uncertainty framework with a fine-tuned ResNet-50 model for surface ordnance classification. Their method used epistemic uncertainty to flag unreliable predictions under noisy, ambiguous, and adversarial inputs. However, since the evaluation used a simulated and controlled dataset, its robustness under real-world domain shifts and OOD deployment conditions remains unresolved. Together, these studies show that RGB-based deep learning models are feasible for visible mine and UXO detection, but their reported performance remains closely tied to the specific datasets, target conditions, and evaluation protocols used in each study.

\subsection{UAV/UGV-Based RGB Datasets for Surface Landmine Detection}
\label{subsec:rgb_datasets_surf_land_det}

UAV/UGV-based RGB public datasets for surface landmine and UXO detection remain limited compared with general remote sensing and object-detection benchmarks. This scarcity is expected because realistic data collection is constrained by safety requirements, restricted access to contaminated or controlled test sites, target availability, and limitations on releasing operational mine-action data. As a result, many studies rely on inert targets, surrogates, replicas, synthetic data, or controlled experimental fields.

\begin{table*}[!t]
\centering
\caption{Existing UAV/UGV-based RGB landmine and UXO datasets relevant to surface-mine detection.}
\label{tab:prior_methods}
\footnotesize
\setlength{\tabcolsep}{2.4pt}
\newcolumntype{L}[1]{>{\raggedright\arraybackslash}p{#1}}
\newcolumntype{Z}{>{\raggedright\arraybackslash}X}
\begin{tabularx}{\textwidth}{@{}
L{0.120\textwidth}
L{0.095\textwidth}
L{0.090\textwidth}
L{0.090\textwidth}
L{0.070\textwidth}
L{0.075\textwidth}
L{0.070\textwidth}
L{0.110\textwidth}
Z
@{}}
\toprule
\textbf{Reference} &
\textbf{Dataset / Study Name } &
\textbf{Target Type} &
\textbf{Platform / Sensor} &
\textbf{RGB Samples} &
\textbf{Labeled Samples} &
\textbf{Target Instances} &
\textbf{IID / OOD Support} &
\textbf{Relevant Characteristics} \\
\midrule

Baur et al.~\cite{rs12050859, baur2021how_to_implement_drones} (2020/2021)
& Custom UAV survey
& PFM-1
& UAV/RGB
& 183 crops used
& 183
& --
& Withheld test settings
& Early UAV RGB detection study; performance decreased under fully withheld testing. \\

\addlinespace
Vivoli et al.~\cite{Vivoli2024DeepImaging} (2024)
& SULAND\_v1
& PFM-1, PMA-2
& Robotic UGV/RGB
& 33{,}771
& 10{,}167
& 10{,}843
& Explicit IID/OOD
& RGB dataset with IID/OOD split and reported OOD degradation. Counts follow the released files used in this study. \\

\addlinespace
Agrawal-Chung and Moin~\cite{agrawalchung2024dronflyby} (2024)
& Drone Flyby
& POM-2, POM-3
& UAV/RGB
& 390 used
& --
& --
& Altitude-based test sets
& Scale models at 2.5, 5, and 10\,m AGL; useful for altitude-dependent detection behavior. \\

\addlinespace
Gallagher and Oughton~\cite{Gallagher2025AMLID:Detection} (2025)
& AMLID
& AP and AT mines
& UAS/RGB + LWIR
& 12{,}078
& 12{,}078
& 14{,}905 (test)
& Multi-condition, no geographic OOD split
& 21 inert simulants across four altitudes, two seasons, three illumination conditions, and 11 RGB--LWIR fusion levels. \\

\addlinespace
Lekhak et al.~\cite{Lekhak2025UncertaintyDropout} (2025)
& Simulated ordnance
& Four simulated ordnance classes
& Simulated RGB
& 5{,}952
& 5{,}952
& --
& Train/val/test; perturbation tests
& Classification dataset used for uncertainty quantification; no bounding-box detection labels or real-scene OOD split. \\

\addlinespace
Malizia et al.~\cite{Malizia2026MineInsight:Environments} (2026)
& MineInsight
& 15 landmine types
& UGV/RGB + VIS - SWIR + LWIR
& $\sim$38{,}000 RGB frames
& --
& --
& Multi-track, no formal RGB IID/OOD split
& Clutter-rich off-road dataset with 15 landmines and 20 distractors; authors note domain gaps across seasons, terrains, and weather. \\

\addlinespace
\textbf{Ours} (SULAND\_v2)
& SULAND\_v2
& PFM-1, PMA-2
& Robotic UGV/RGB
& 33{,}771
& 11{,}560
& 12{,}433
& Explicit IID/OOD
& Refined SULAND annotations with corrected class convention; supports v1/v2 validation and broad detector benchmarking under domain shift. \\

\bottomrule
\vspace{0.05in}
\end{tabularx}
\begin{minipage}{0.98\textwidth}
\footnotesize
A dash indicates that a directly comparable count was not reported or is not applicable.
\end{minipage}
\end{table*}

Several datasets have nevertheless been introduced or used to support RGB-based mine and UXO detection research, although their scope and evaluation settings vary considerably. Table~\ref{tab:prior_methods} summarizes existing UAV/UGV-based RGB datasets relevant to surface-mine detection, highlighting differences in target type, acquisition platform, sensing modality, sample size, annotation availability, and IID/OOD support. Baur \textit{et al.}~\cite{baur2021how_to_implement_drones, rs12050859} used UAV-acquired RGB imagery for PFM-1 detection and evaluated performance under partially and completely withheld test settings. Their dataset and evaluation protocol provided an early demonstration of drone-based RGB detection for surface mines, but the reported performance drop under completely withheld testing suggests sensitivity to changes in scene composition, target visibility, and acquisition conditions. Agrawal-Chung and Moin~\cite{agrawalchung2024dronflyby} introduced a drone-based RGB dataset collected over test sites seeded with physical scale models of Russian POM-2 and POM-3 surface mines. Although the dataset supports evaluation across three deployment altitudes, making it useful for studying altitude-dependent detection behavior, its small scale limits extensive benchmarking. Gallagher and Oughton released the AMLID dataset~\cite{Gallagher2025AMLID:Detection}, which includes 21 inert landmine simulants captured across four operational altitudes, two seasons, and three illumination conditions. This provides a valuable framework for controlled in-distribution evaluation under varying acquisition conditions; however, it does not fully address OOD settings involving unseen geographic backgrounds, diverse soil compositions, or modified target appearances. Malizia\textit{ et al}. introduced MineInsight~\cite{Malizia2026MineInsight:Environments}, a multimodal dataset that includes RGB imagery for 15 different mine types. While MineInsight broadens the range of mine categories and sensing modalities available for study, the authors note that domain gaps remain when transferring across different seasons, terrains, and weather conditions, including sunny, snowy, arid, and grassy environments. Similarly, the simulated dataset used by Lekhak \textit{et al.}~\cite{Lekhak2025UncertaintyDropout} supports controlled evaluation of uncertainty-aware ordnance classification, but it does not capture the full variability of real-world deployment conditions. Overall, existing datasets are valuable for advancing RGB-based mine and UXO detection, but they remain limited for systematic evaluation of domain generalization across unseen locations, soil types, backgrounds, target appearances, and acquisition conditions.

\subsection{Domain Shift and Benchmarking Gaps in RGB-Based Surface Mine Detection }
\label{benchmark_limitations_and_domain_shift}

Although prior studies have demonstrated the feasibility of applying deep learning models to RGB-based mine and UXO detection, existing evaluations remain limited from both model and dataset perspectives. On the model side, recent years have introduced a wide range of object-detection architectures, including two-stage detectors, one-stage detectors, transformer-based detectors, and open-vocabulary detection models \cite{yolov8, khanam2024yolov11, tian2026yolov12, sapkota2025yolo26,  cheng2024yoloworld, zhao2024detrs, 2015_FasterRCNN, peng2024dfine, robinson2025rf}. However, despite the rapid development of modern object-detection architectures, only a limited subset has been applied to surface mine detection \cite{ baur2021how_to_implement_drones, joint_fusion_and_detection_2023, rs12050859, agrawalchung2024dronflyby, Lekhak2025UncertaintyDropout}, and even fewer studies have evaluated detector performance under explicit OOD or domain-shift settings \cite{Vivoli2024DeepImaging}. As a result, it remains unclear \textit{how different detector families behave when the deployment environment differs from the training environment.}

On the dataset side, existing RGB mine and UXO datasets seem to differ in target type, acquisition platform, altitude, background, annotation format, and evaluation protocol \cite{ baur2021how_to_implement_drones, Vivoli2024DeepImaging,rs12050859, agrawalchung2024dronflyby, Lekhak2025UncertaintyDropout,Gallagher2025AMLID:Detection,  Malizia2026MineInsight:Environments}. Many are useful for controlled or in-distribution evaluation, but they are not designed to systematically test generalization to shifted deployment conditions. This is a critical limitation for surface mine detection because detectors may perform well on familiar scenes but fail under new geographic locations, soil types, illumination conditions, target appearances, or cluttered backgrounds. Therefore, reported performance across studies does not necessarily reflect the relative robustness of different detectors under realistic domain shift.

In UAV- and UGV-based surface landmine detection, domain shift can arise from changes in soil texture, vegetation, illumination, shadows, target scale, camera viewpoint, platform motion, terrain slope, occlusion, season, and geographic location. These factors directly affect the visual separability of small mine-like objects from background clutter and can cause detectors to rely on scene-specific correlations rather than target-relevant cues. In safety-critical applications such as humanitarian demining, where missed detections can have severe consequences and clearance remains slow, hazardous, and resource-intensive, OOD evaluation is essential for determining whether a detector has learned features that generalize beyond the training environment.

However, because many existing datasets do not provide explicit IID/OOD split settings or standardized domain-generalization protocols, it remains difficult to compare detection algorithms fairly or identify which models remain robust in unseen environments. These limitations motivate the present work, which adopts SULAND\_v1 as the starting point, refines its annotations to construct SULAND\_v2, and evaluates modern object-detection algorithms under standardized IID and OOD settings.

\section{SULAND\_v1: Dataset Selection, Characteristics, and Benchmark Limitations}
 \label{sec:suland_v1}

\subsection{Rationale for SULAND\_v1 Selection for OOD Analysis}
\label{subsec:suland_v1_ood analysis}

\begin{figure}[t]
    \centering
    \begin{subfigure}[b]{0.22\textwidth}
        \centering
        \includegraphics[width=\textwidth,height=2.8cm,keepaspectratio]{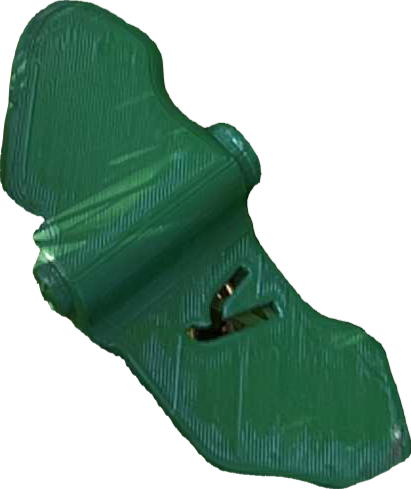}
        \caption{PFM-1}
        \label{fig:pfm1}
    \end{subfigure}
    \hfill
    \begin{subfigure}[b]{0.22\textwidth}
        \centering
        \includegraphics[width=\textwidth,height=2.8cm,keepaspectratio]{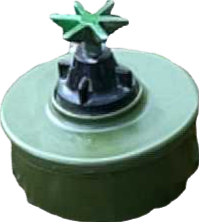}
        \caption{PMA-2}
        \label{fig:pma2}
    \end{subfigure}
    \caption{Representative target class examples segmented from the SULAND\_v1 dataset \cite{Vivoli2024DeepImaging} : (a) PFM-1 ``Butterfly'' mine and (b) PMA-2 ``Starfish'' mine.}
    \label{fig:landmine_types}
\end{figure}

SULAND\_v1 is a publicly released RGB image dataset for surface landmine detection introduced by Vivoli\textit{ et al.}~\cite{Vivoli2024DeepImaging}. The dataset focuses on two target classes: PFM-1, commonly referred to as the ``Butterfly'' mine, and PMA-2, commonly referred to as the ``Starfish'' mine. Representative examples of the two target types are shown in Fig.~\ref{fig:landmine_types}. The dataset is designed to support real-time optical detection of surface landmine surrogates and to evaluate detector behavior under both familiar and shifted environmental conditions. A key feature of SULAND\_v1 is its separation between in-distribution (IID) and out-of-distribution (OOD) data. The IID subset contains PFM-1 and PMA-2 targets imaged in Italy over grass and gravel surfaces under sunny, cloudy, and shadowed conditions, with additional scene clutter such as bushes, branches, walls, bars, tree trunks, and rocks. In contrast, the OOD subset introduces a geographic domain shift from Italy to the United States and includes additional variation in environmental context, vegetation, slope, camera viewpoint, target distance, partial occlusion, target size, color, and appearance. This structure makes SULAND\_v1 a useful starting point for studying RGB-based surface landmine detection and for evaluating which detectors remain robust under environmental and geographic changes.

\begin{figure}[!t]
  \centering
  \includegraphics[width=0.98\linewidth]{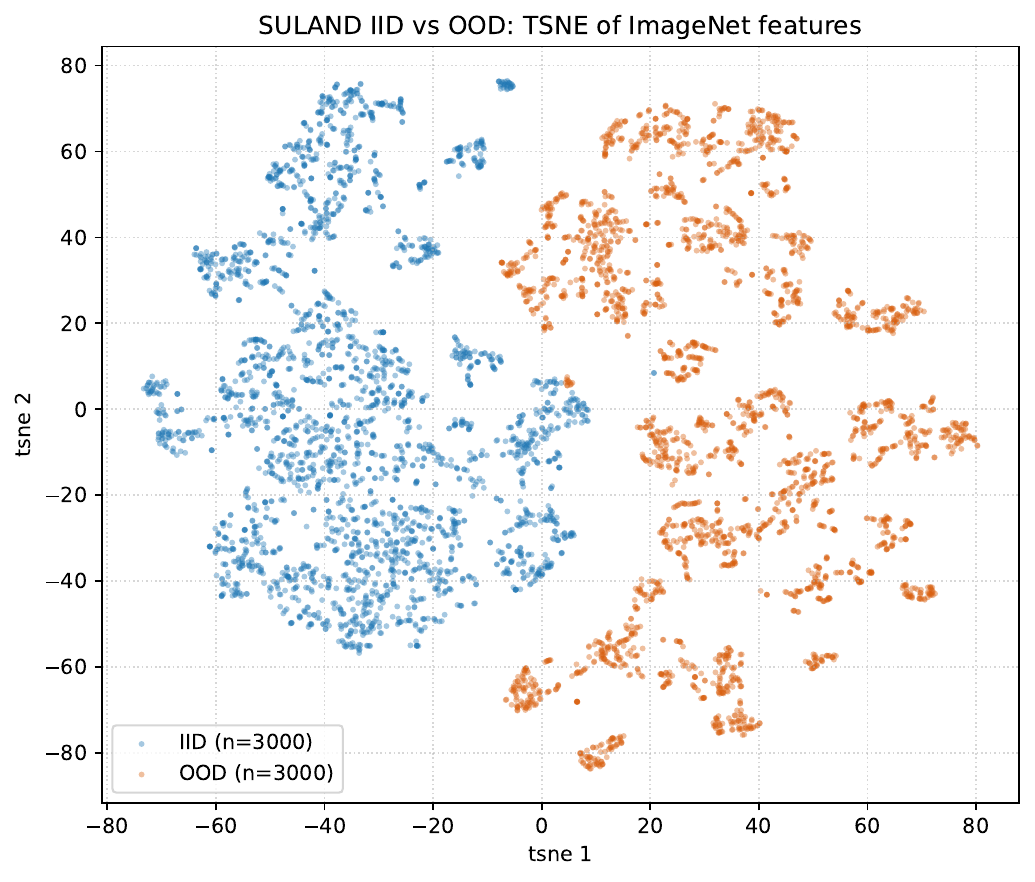}
  \caption{t-SNE projection of ImageNet-pretrained features for 3000 randomly selected IID samples and 3000 randomly selected OOD samples from SULAND\_v1, illustrating feature-space separation between the two distributions.}
\label{fig:imagenet_tsne_iid_ood}
\end{figure}

Although the SULAND\_v1 study defined the IID and OOD subsets primarily based on visual differences and qualitatively reported degraded YOLOv8 performance on the OOD data, quantitative OOD metrics and detailed experimental settings were not provided. To independently assess the proposed IID/OOD separation, we extracted feature embeddings using an ImageNet-pretrained backbone \cite{5206848} and projected them with t-SNE. Using a generic pretrained backbone provides a detector-independent view of the visual distribution shift, separate from the object-detection models evaluated later in this study. As shown in Fig.~\ref{fig:imagenet_tsne_iid_ood}, 3000 randomly selected samples from each subset form largely separated regions in feature space, providing qualitative evidence of visual distribution shift and motivating explicit IID/OOD benchmarking for detector robustness.

\subsection{SULAND\_v1 Dataset Organization and Need for Refinement}
\label{subsec:suland_v1_refinement}

The original SULAND\_v1 release is an important contribution because it provides one of the few publicly available RGB datasets for surface landmine detection with an explicit OOD evaluation setting. Our audit is therefore not intended to diminish its value, but to assess whether the released annotations are sufficiently consistent for reliable benchmarking and model comparison.

SULAND\_v1 is formulated as a two-class object-detection dataset with bounding-box annotations for PFM-1 and PMA-2 targets. The dataset was constructed by extracting frames from multiple video sequences and annotating visible target instances in the resulting images. The released dataset is organized into IID and OOD subsets following a standard object-detection format with separate image and label folders. In the released archive available to the authors of this paper, the IID subset contains 45 folders in total, consisting of 34 training folders, 5 validation folders, and 6 test folders. The OOD subset contains 10 evaluation folders, denoted as US1--US10, and is included as an additional evaluation split under the validation directory.

During dataset inspection, we found that this released folder structure differs from the sequence counts reported in the original SULAND\_v1 publication \cite{Vivoli2024DeepImaging}, which lists 47 in-distribution and 11 out-of-distribution video sequences. Therefore, all dataset statistics, annotation audits, and benchmark experiments in this study are based on the released files available to us rather than the aggregate counts reported in the original publication.

\begin{table}[!t]
\centering
\caption{SULAND dataset statistics before (v1) and after (v2) re-annotation and cleaning.}
\label{tab:v1v2_stats}
\resizebox{0.48\textwidth}{!}{%
\begin{tabular}{l l l l l l}
\toprule
\textbf{Ver.} & \textbf{Split} & \textbf{Images} & \textbf{Foreground} & \textbf{Background} & \textbf{Annotations} \\
\midrule
v1 & IID Train & 22,756 & 5,234  & 17,522 & 5,510 \\
v2 & IID Train & 22,756 & 6,272  & 16,484 & 6,714 \\
\addlinespace
v1 & IID Val   &  2,836 &   567  &  2,269 &   654 \\
v2 & IID Val   &  2,836 &   696  &  2,140 &   807 \\
\addlinespace
v1 & IID Test  &  3,743 &   839  &  2,904 &   892 \\
v2 & IID Test  &  3,743 & 1,006  &  2,737 & 1,056 \\
\addlinespace
v1 & OOD Val   &  4,436 & 3,527  &    909 & 3,787 \\
v2 & OOD Val   &  4,436 & 3,586  &    850 & 3,856 \\
\bottomrule
\end{tabular}}
\end{table}

Our audit showed that SULAND\_v1 is moderately imbalanced and contains several annotation inconsistencies. Table~\ref{tab:v1v2_stats} summarizes the foreground/background composition and annotation counts before and after refinement. The refinement procedure and quantitative changes between SULAND\_v1 and SULAND\_v2 are further discussed in detail in Section~\ref{subsec:quantitative_changes_v1_v2}. In the Table~\ref{tab:v1v2_stats}, \emph{foreground} denotes an image with at least one target annotation, \emph{background} an image with none, and \emph{annotations} the total number of bounding-box instances; annotations may exceed foreground images because an image can contain multiple targets. In the original IID training, validation, and test splits, foreground images represent only about 20--23\% of each split. For example, in the IID training split, 5{,}234 of 22{,}756 images contain at least one annotation, while 17{,}522 images are labeled as background-only. In contrast, approximately 79.5\% of the OOD images contain at least one target annotation, showing that the IID and OOD subsets differ not only in environmental conditions but also in foreground/background composition. In addition to this imbalance, our audit identified missing annotations, missing label files, false annotations, mislocalized bounding boxes, inconsistent treatment of partially visible targets, class-label inconsistencies, and non-representative artifacts attached to or near targets.

The foreground/background imbalance is not inherently problematic. Survey-style imagery naturally contains many frames without visible targets, and background-only images can help detectors learn diverse non-target appearances and reduce false positives in cluttered outdoor scenes. However, this benefit depends on annotation completeness and consistency. Background-only frames are useful only when they are truly free of visible target objects; otherwise, visible but unannotated targets are incorrectly treated as background during training and evaluation.

The identified annotation issues can affect both model learning and benchmark reliability. Missing or incorrect annotations introduce conflicting supervision, mislocalized boxes reduce the reliability of localization-sensitive metrics, and inconsistent annotation criteria can cause visually similar samples to receive different training or evaluation labels. This issue is particularly important for SULAND\_v1 because the dataset is derived from video sequences, where adjacent frames may contain similar target appearances but receive different annotation decisions. In a scarce-data and safety-critical setting such as surface mine detection, these inconsistencies can lead to misleading conclusions about detector accuracy, OOD robustness, and the relative value of different model families.

The following subsection presents the major annotation issues identified in SULAND\_v1 and explains how they motivated the construction of SULAND\_v2 as a refined benchmark for standardized IID/OOD evaluation.

\subsection{Audit Protocol and Annotation Issues in SULAND\_v1 Dataset}
\label{subsec:annotation_issues_in_Suland_v1}
 \begin{figure*}[!t]
\centering
\begin{subfigure}[t]{0.33\textwidth}
    \centering
    \includegraphics[width=\linewidth]{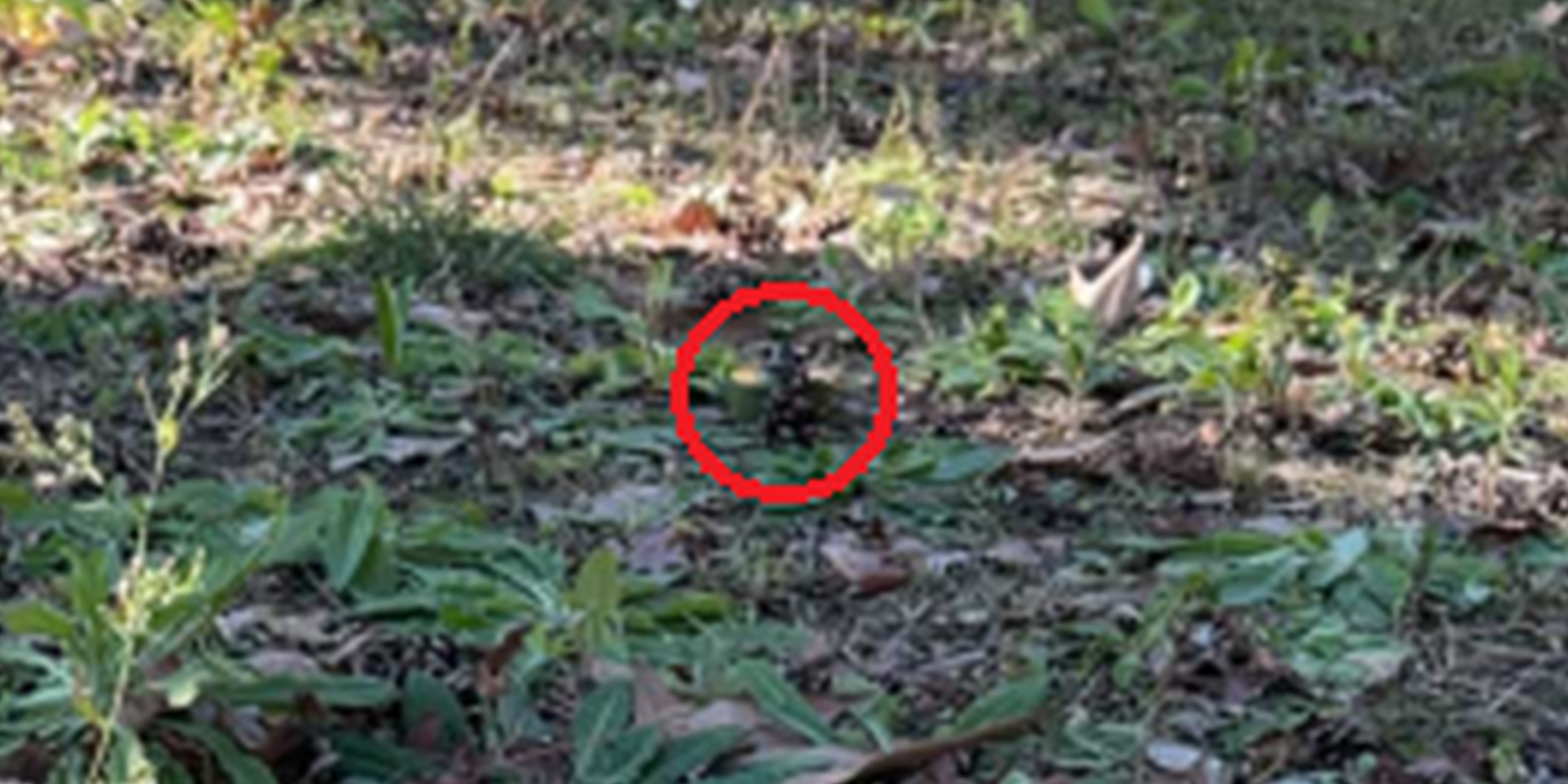}
    \caption{Clearly visible target unannotated (sample \texttt{ITA-v17-95}).}
    \label{fig:img1}
\end{subfigure}\hfill
\begin{subfigure}[t]{0.33\textwidth}
    \centering
    \includegraphics[width=\linewidth]{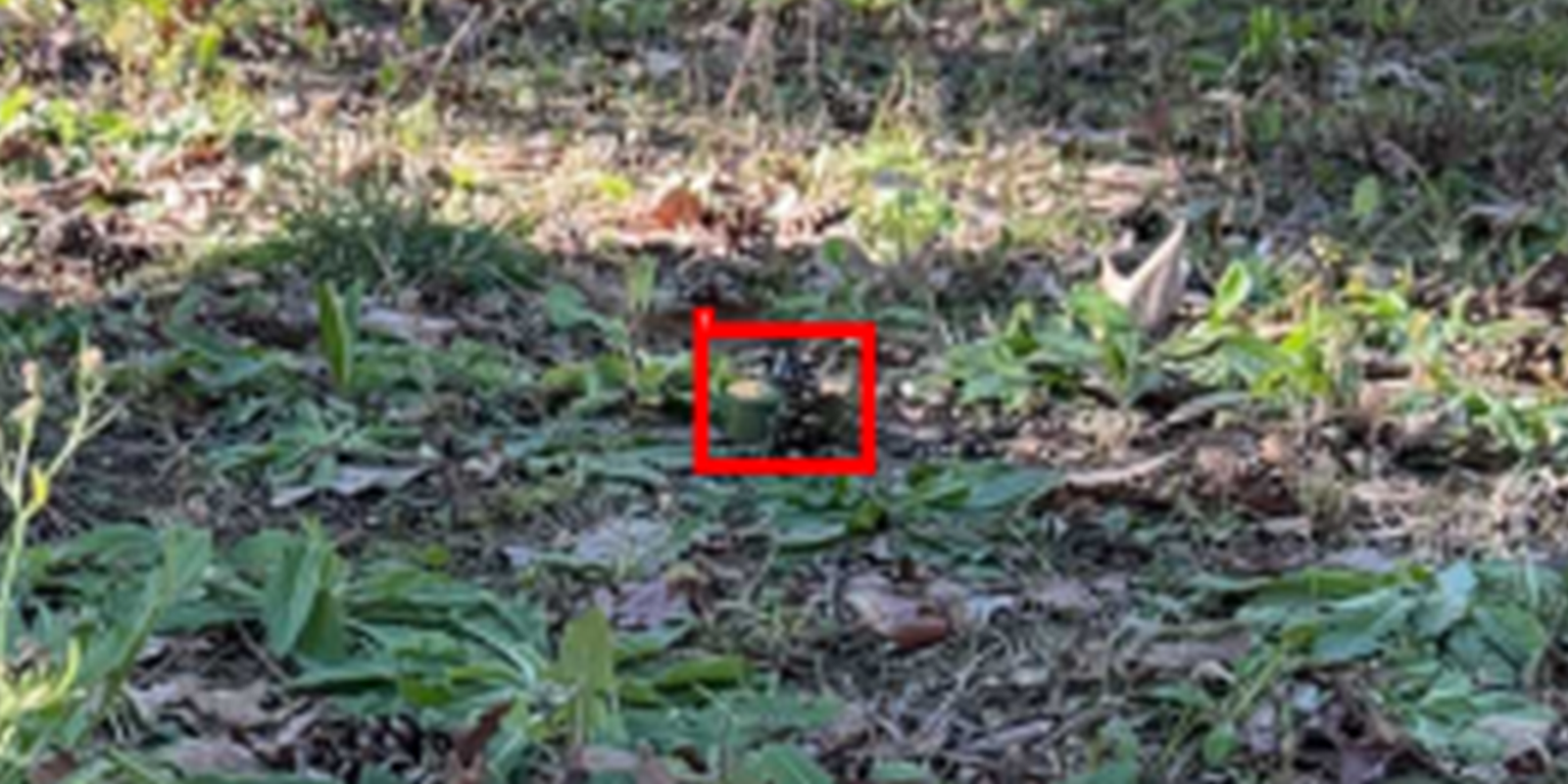}
    \caption{Visually similar target annotated (sample  \texttt{ITA-v17-99}).}
    \label{fig:img2}
\end{subfigure}\hfill
\begin{subfigure}[t]{0.32\textwidth}
    \centering
    \includegraphics[width=\linewidth]{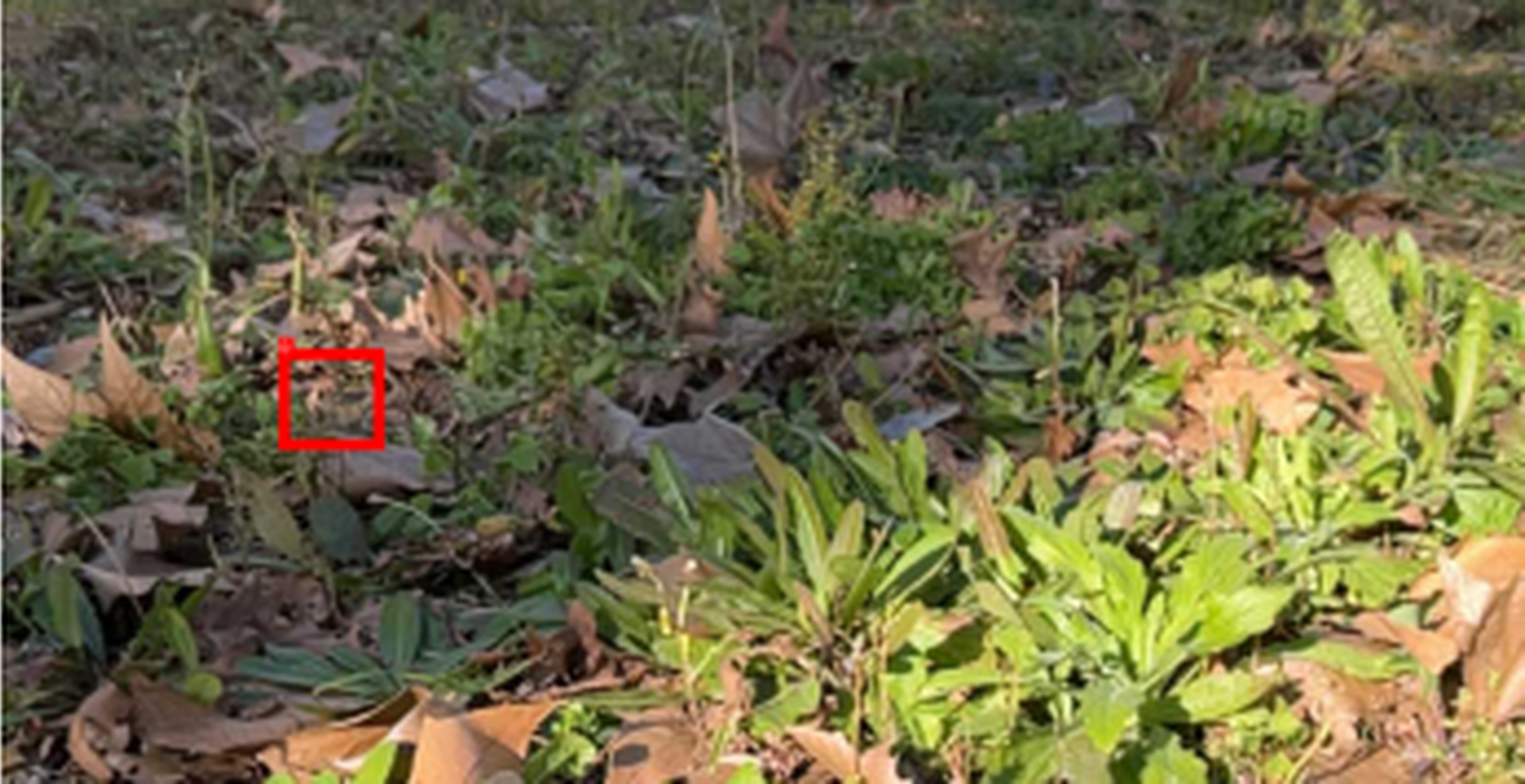}
    \caption{Annotation propagation without any targets in scene (several samples in  \texttt{ITA-v14}).}
    \label{fig:img3}
\end{subfigure}

\vspace{0.6em}

\begin{subfigure}[t]{0.322\textwidth}
    \centering
    \includegraphics[width=\linewidth]{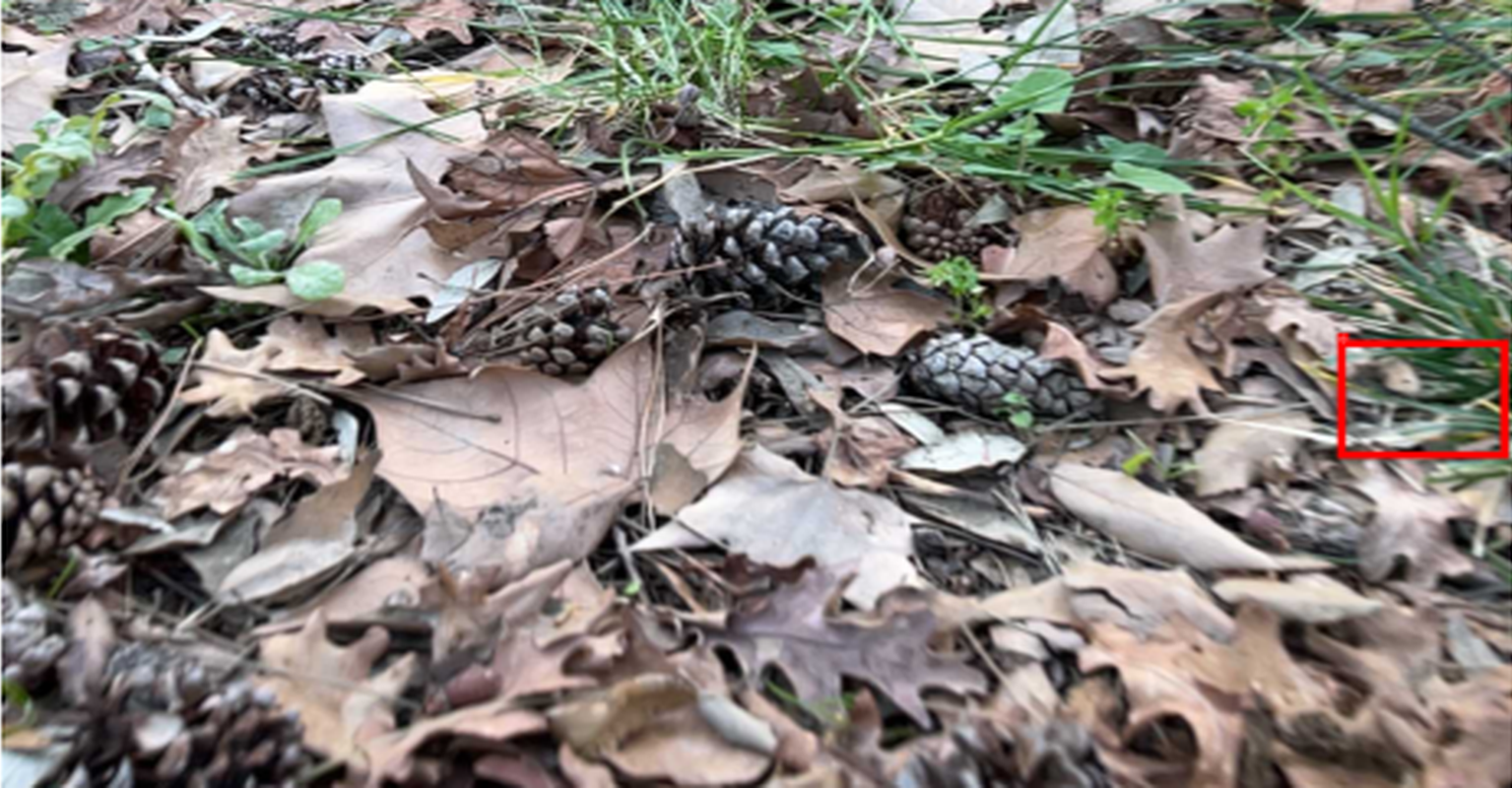}
    \caption{False positive annotation with no targets in scene (sample  \texttt{ITA-v1-43}).}
    \label{fig:img4}
\end{subfigure}\hfill
\begin{subfigure}[t]{0.335\textwidth}
    \centering
    \includegraphics[width=\linewidth]{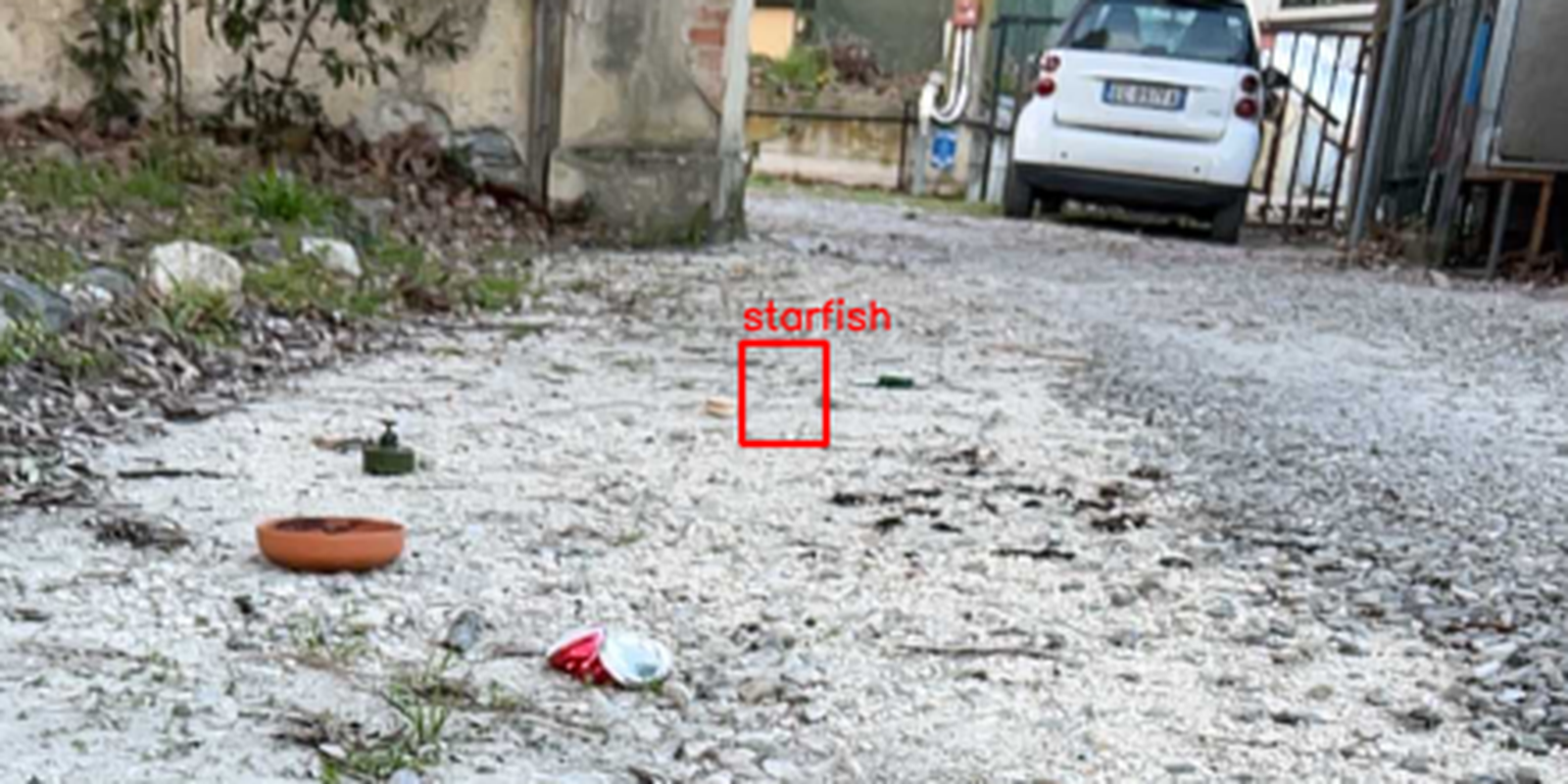}
    \caption{Misaligned bounding box (sample \texttt{ITA-v4-329}).}
    \label{fig:img5}
\end{subfigure}\hfill
\begin{subfigure}[t]{0.32\textwidth}
    \centering
    \includegraphics[width=\linewidth]{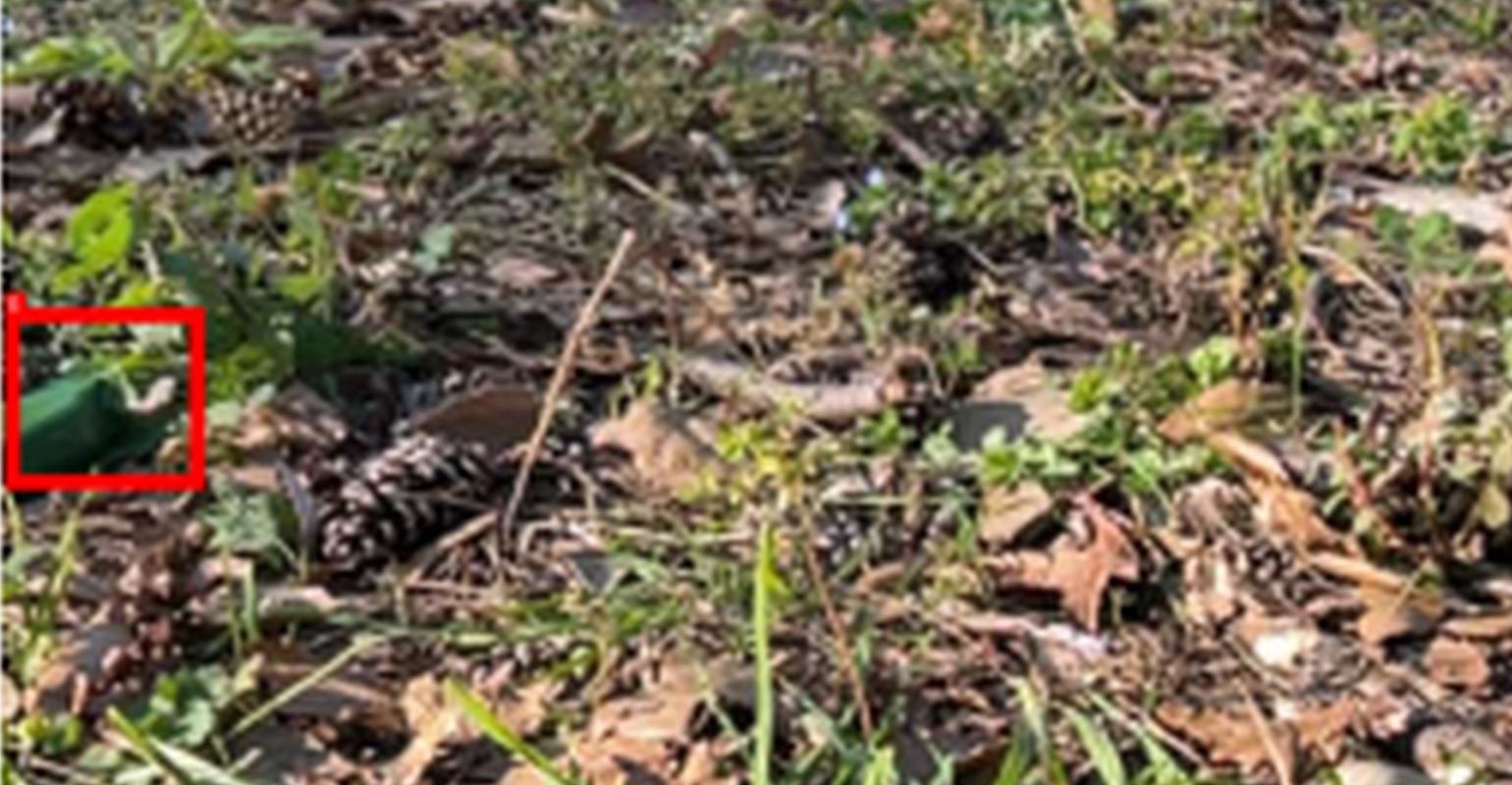}
    \caption{Annotated partially visible target (sample \texttt{ITA-v18-235})}
    \label{fig:img6}
\end{subfigure}

\vspace{0.6em}

\begin{subfigure}[t]{0.32\textwidth}
    \centering
    \includegraphics[width=\linewidth]{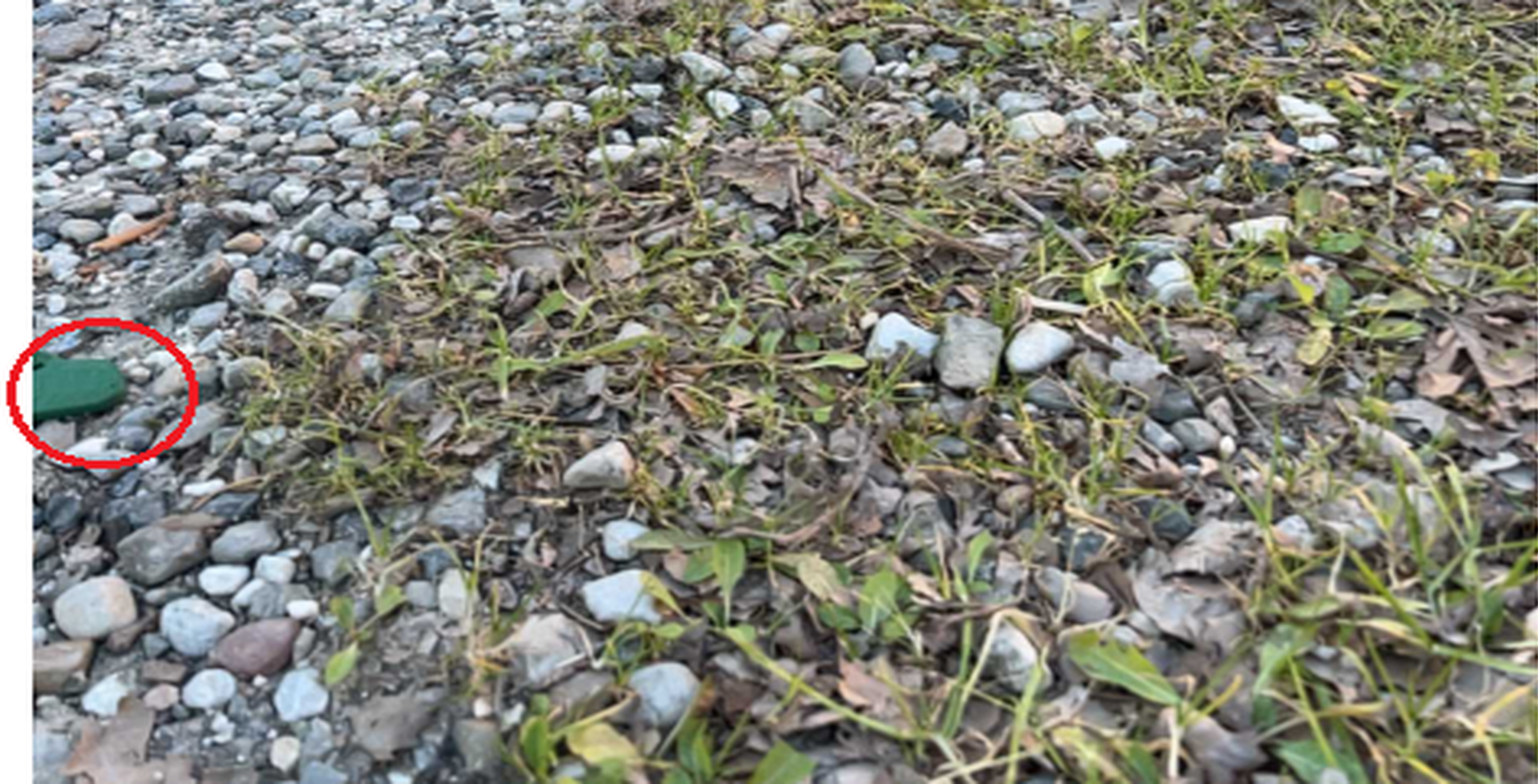}
    \caption{Comparable partially visible target unannotated (sample \texttt{ITA-v16-61}).}
    \label{fig:img7}
\end{subfigure}\hfill
\begin{subfigure}[t]{0.32\textwidth}
    \centering
    \includegraphics[width=\linewidth]{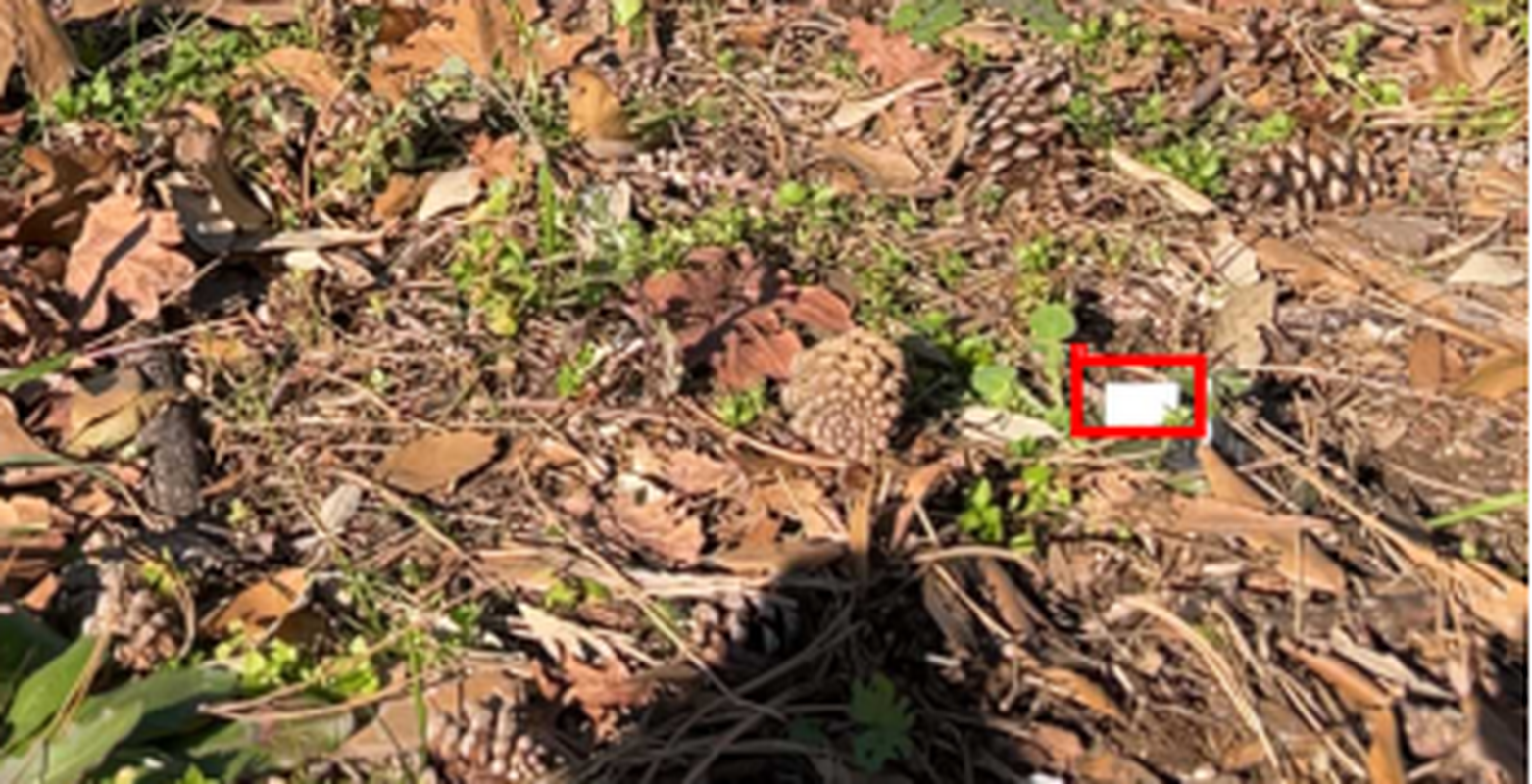}
    \caption{Bounding box enclosing an artificial marker instead of nearby PMA-2 target (sample \texttt{ITA-v14-32}).}
    \label{fig:img8}
\end{subfigure}\hfill
\begin{subfigure}[t]{0.33\textwidth}
    \centering
    \includegraphics[width=\linewidth]{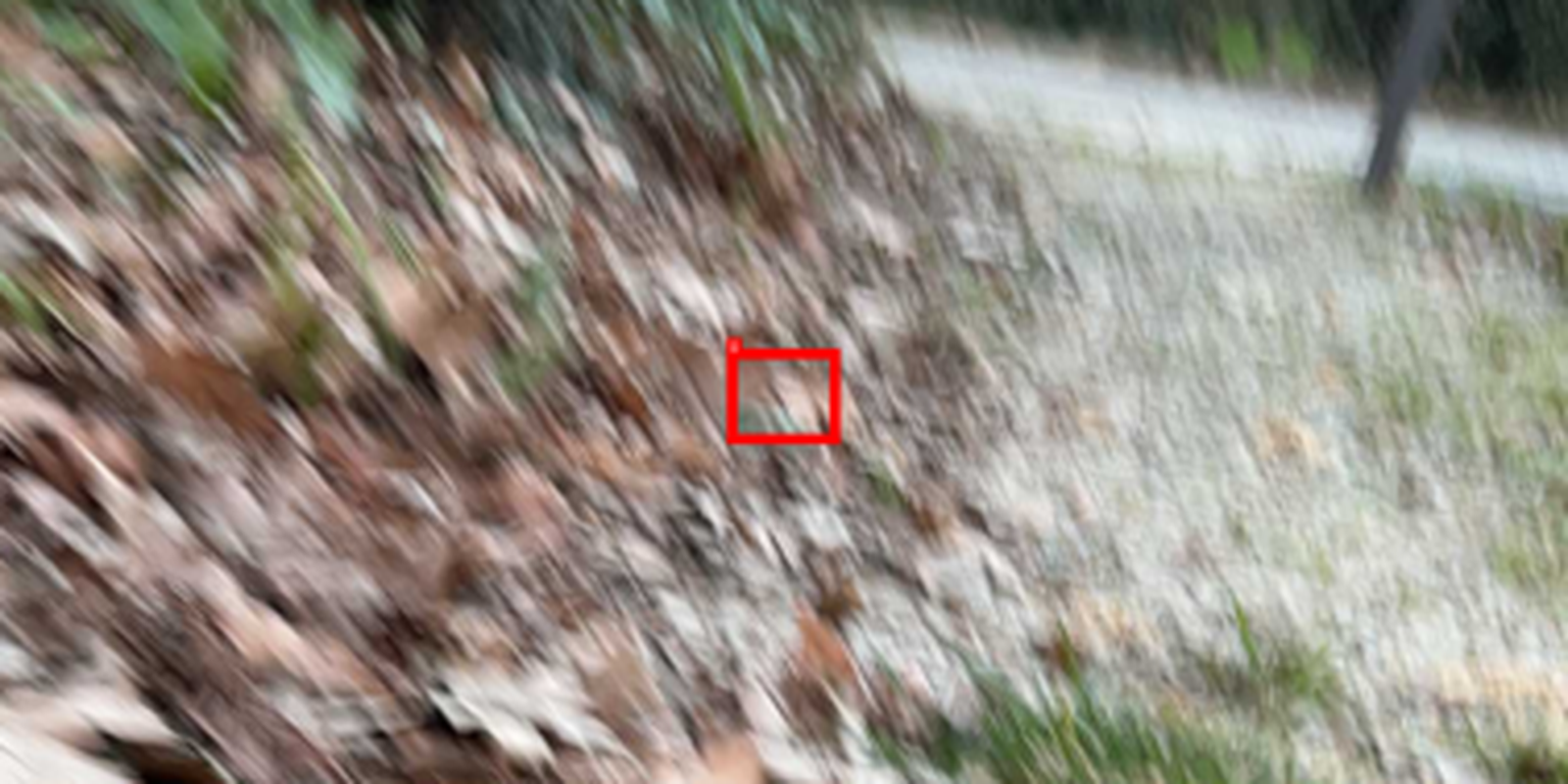}
    \caption{Severe motion blur and resulting annotation uncertainty (sample \texttt{ITA-v30-707}).}
    \label{fig:img9}
\end{subfigure}

\caption{Representative cropped examples of annotation defects identified during the SULAND\_v1 dataset audit. While the audit establishes seven error categories, six are illustrated here: missing or incomplete annotations (a–b), false positive and propagation errors (c–d), mislocalized flaws (e), inconsistent partial-visibility criteria (f–g), non-representative artifacts (h), and quality degradation (i).}
\label{fig:dataset_problems}

\end{figure*}

To assess the reliability of SULAND\_v1, we conducted a systematic manual audit of the released images and annotations. Each sample was inspected folder-by-folder and frame-by-frame using a custom Python visualization script. For each image, the script loaded the corresponding YOLO-format label file and overlaid the author-provided bounding boxes on the image, allowing the visible image content and released annotation to be evaluated simultaneously. When needed, the displayed bounding boxes were cross-checked against the coordinate values in the corresponding label files to verify target localization.

The audit was performed in two passes. In the first pass, each sample was manually visualized and recorded in an audit spreadsheet as either requiring correction or not requiring correction. For samples requiring correction, we recorded the observed issue and supporting notes, including the target class, error type, approximate frame or sample range, object visibility, and whether the issue appeared as an isolated error or as part of a temporal sequence. Since SULAND\_v1 was derived from video sequences, this frame-by-frame review allowed us to follow target appearances over time and identify annotation inconsistencies across adjacent or nearby frames. The complete audit spreadsheet is provided as supplementary material and will be released with the SULAND\_v2 dataset repository.

\begin{figure}[!t]
  \centering
  \includegraphics[width=0.98\linewidth]{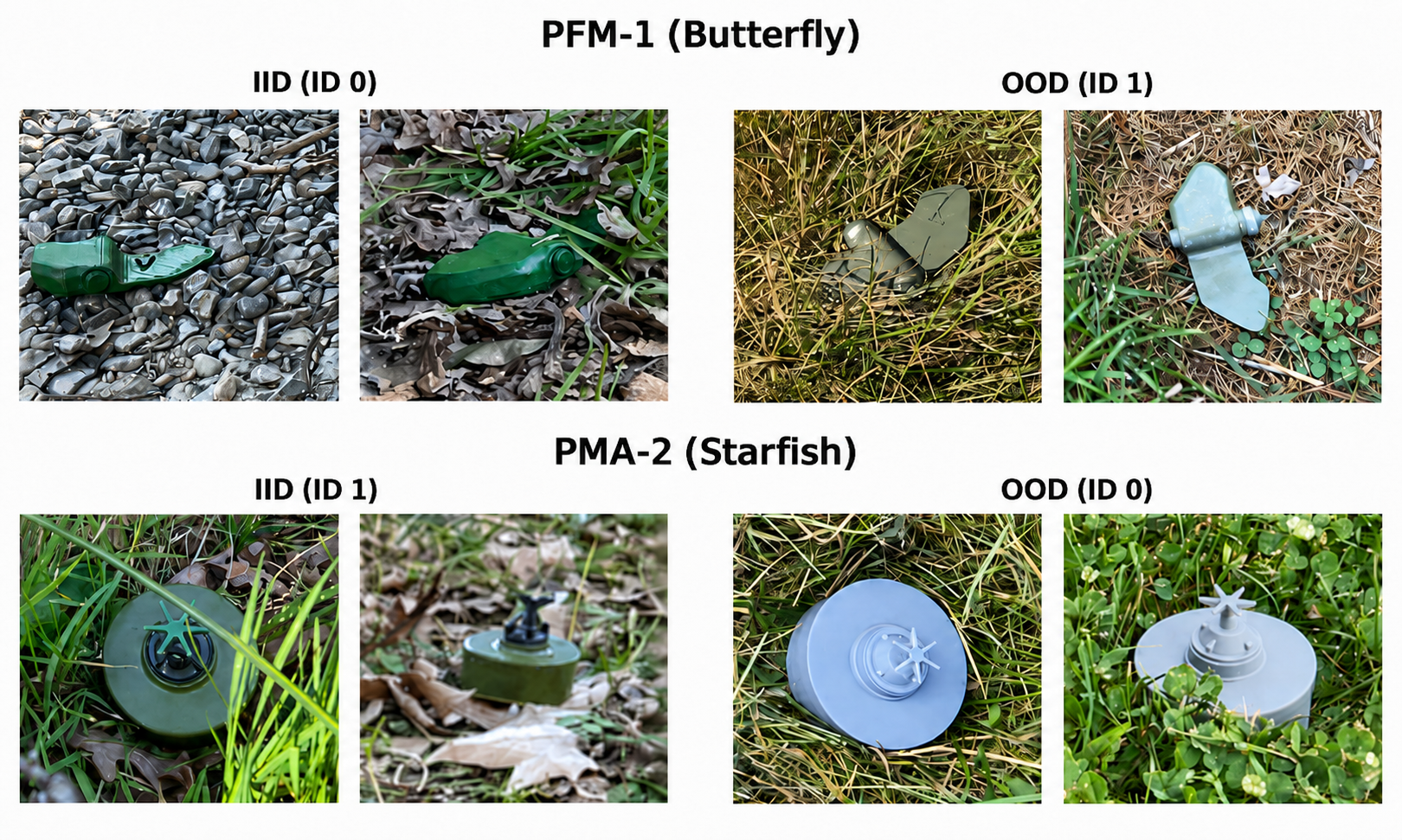}
  \caption{Representative examples of class-ID mismatch in SULAND\_v1. The IID subset assigns PFM-1 and PMA-2 to class IDs 0 and 1, respectively, while the OOD subset uses the reversed mapping.}
\label{fig:class_id_mismatch}

\end{figure}

During the second pass, all identified defective samples were systematically reviewed and mapped to seven major annotation-error categories. Six of these categories are visually illustrated through representative cropped examples in Fig.~\ref{fig:dataset_problems}, while the seventh error category is presented separately in Fig.~\ref{fig:class_id_mismatch}. Additional folder-level details are provided in the supplementary material.
The seven distinct error categories are as follows:
\begin{itemize}
    \item \textbf{Missing or Incomplete Annotations:} Visible targets that completely lack bounding boxes, often due to delayed annotation onset within video sequences (illustrated in Fig.~\ref{fig:dataset_problems}(a) and (b)). In some samples, the corresponding \texttt{.txt} label files were also entirely missing.
    \item \textbf{False Positive and Propagation Errors:} Bounding boxes that persist across empty frames after a target has exited the scene, or empty background regions incorrectly labeled as mines (illustrated in Fig.~\ref{fig:dataset_problems}(c) and (d)).
    \item \textbf{Mislocalized Flaws:} Bounding boxes that are significantly shifted, misaligned, or poorly localized relative to the true target boundaries (illustrated in Fig.~\ref{fig:dataset_problems}(e)).
    \item \textbf{Inconsistent Partial-Visibility Criteria:} Ambiguous annotation thresholds where an occluded or partially visible target is labeled in one frame but left unannotated in a visually comparable frame (illustrated in Fig.~\ref{fig:dataset_problems}(f) and (g)).
    \item \textbf{Non-Representative Artifacts:} Bounding boxes that mistakenly enclose artificial ground markers or physical tags rather than focusing strictly on the target object (illustrated in Fig.~\ref{fig:dataset_problems}(h)).
    \item \textbf{Quality Degradation:} Instances where severe environmental or motion blur obscures target boundaries, drastically increasing annotation uncertainty (illustrated in Fig.~\ref{fig:dataset_problems}(i)).
    \item \textbf{Class-ID Mismatch:} Class-ID mismatch between the IID and OOD subsets, as shown in Fig.~\ref{fig:class_id_mismatch}. The IID subset encodes PFM-1 and PMA-2 as class IDs 0 and 1, respectively, whereas the OOD subset uses the reversed mapping. The effect of this issue on benchmark performance is examined in detail in Section~\ref{subsec:v1_v2_cross_evaluation_protocol}.
\end{itemize}

The audit revealed a large number of samples that need correction in the dataset. Therefore, rather than applying only isolated local edits, we manually re-annotated the whole dataset sample-by-sample using Label Studio~\cite{Label_Studio}. The resulting corrected annotations define SULAND\_v2 and are used for all benchmark experiments in this study.

\section{SULAND\_v2: A Refined Version for Benchmarking}
\label{sec:suland_v2}

\subsection{Re-Annotation Principles and Correction Procedure}
\label{subsec:reannotation_procedure}

SULAND\_v2 preserves the original image collection, two-class detection task, and IID/OOD split structure of SULAND\_v1, while correcting annotation inconsistencies identified during the audit. In the refined annotations, Class~0 corresponds to PFM-1 and Class~1 corresponds to PMA-2 across all splits. The objective of SULAND\_v2 is therefore not to alter the dataset content, but to provide a more consistent annotation set for reliable IID/OOD benchmarking.

Each sample was reopened in Label Studio~\cite{Label_Studio} and manually re-annotated using the audit records described in Section~\ref{subsec:annotation_issues_in_Suland_v1} as guidance. The original images and filenames were preserved unchanged; only the annotation files were revised. Missing or incomplete annotations were added, invalid annotations were removed, inaccurate boxes were redrawn, and class-label inconsistencies were corrected. All revisions were performed through sample-by-sample manual review rather than automated correction.

A target was considered eligible for annotation when a visible and class-identifiable portion of the object could be discerned. Since SULAND\_v1 was extracted from video sequences, uncertain cases were not judged in isolation. Adjacent frames were examined when needed to verify target presence, class identity, and temporal continuity. This temporal context was used only to resolve target identity and visibility, not to annotate fully occluded or non-discernible targets.

Partially visible targets were annotated when the visible region remained class-identifiable. This criterion was applied consistently across consecutive frames and folders because operational imagery may contain targets partially occluded by vegetation, terrain, debris, shadows, or image boundaries. Fully occluded, severely blurred, or otherwise non-discernible targets were left unannotated unless their class and visible extent could be established with sufficient confidence.

Samples containing stickers, flags, markers, or other non-representative artifacts were not automatically excluded. When a discernible portion of the mine body remained visible, the sample was retained and the bounding box was restricted to the visible mine region. Annotations were removed only when the labeled region primarily covered an artifact and did not provide sufficient visual evidence of the target.

Bounding boxes were drawn tightly around the visible target extent. This is important because object-detection evaluation depends on localization quality through IoU-based matching. Oversized boxes can include unnecessary background and encourage reliance on surrounding context, whereas undersized or displaced boxes can omit discriminative target features and introduce localization inconsistency. For partially occluded targets, the box enclosed only the visible target region rather than an inferred full-object boundary.

After the initial re-annotation, the revised labels passed through a structured quality-control stage to check temporal, spatial, and class consistency. Each SULAND\_v2 annotation was compared with its SULAND\_v1 counterpart at the bounding-box level, and each box-level change was assigned to one of four correction scenarios: \emph{added}, \emph{removed}, \emph{tightened}, or \emph{class-corrected}. An \emph{added} box refers to a target annotation present only in SULAND\_v2. A \emph{removed} box refers to an annotation present in SULAND\_v1 but removed in SULAND\_v2 because no valid target was visible. A \emph{tightened} box refers to a same-class annotation whose geometry was refined to better match the visible target extent. A \emph{class-corrected} box refers to a matched annotation whose class label was changed between PFM-1 and PMA-2. The quantitative distribution of these correction types is reported in the following subsection.

To make exhaustive verification tractable, each changed box was rendered as a local crop centered on the correction and tiled into contact-sheet pages containing 50 crops. This allowed corrections of the same type to be reviewed efficiently while preserving enough local context to judge annotation validity. Three reviewers independently inspected the contact sheets for each correction scenario, and each correction was accepted by majority vote. The final annotations were exported in YOLO format to maintain compatibility with the original dataset structure and were later converted to COCO format for benchmark experiments.

\subsection{Quantitative Comparison of SULAND\_v1 and SULAND\_v2 and Representative Annotation Corrections}
\label{subsec:quantitative_changes_v1_v2}

 \begin{table*}[!t]
  \centering
  \caption{Split-level annotation changes from SULAND\_v1 to SULAND\_v2.}
  \label{tab:annotation_revision}
  \begin{tabular}{l l l l l l l l l l l}

  \toprule
  \textbf{Split} & \textbf{Images} & \textbf{Inst.\ (v1$\to$v2)} &
  \textbf{Added} & \textbf{Removed} & \textbf{Tightened} & \textbf{Class-fix} &
  \textbf{Med.\ IoU} & \textbf{Med.\ ratio} &
  \textbf{\%IoU$<$0.7} & \textbf{\%shrunk} \\
  \midrule
  IID train & 22{,}756 & 5{,}510 $\to$ 6{,}714 & 3{,}134 & 1{,}930 & 3{,}580 & 0 & 0.72 & 0.85 & 44.5 & 77.9 \\
  IID val   &  2{,}836 &   654 $\to$   807 &   490 &   337 &   317 & 0 & 0.73 & 0.90 & 40.4 & 69.4 \\
  IID test  &  3{,}743 &   892 $\to$ 1{,}056 &   568 &   404 &   488 & 0 & 0.72 & 0.91 & 44.9 & 68.0 \\
  OOD val   &  4{,}436 & 3{,}787 $\to$ 3{,}856 &   693 &   624 & 3{,}163 & 3{,}162 & 0.82 & 0.97 & 19.4 & 57.7 \\
  \bottomrule
  \end{tabular}
\end{table*}

SULAND\_v1 and SULAND\_v2 contain the same image files and splits; therefore, the differences reported in Table~\ref{tab:v1v2_stats} result entirely from annotation refinement. The table provides a split-level summary of the changes in foreground images, background-only images, and bounding-box annotations. Overall, the number of annotated instances increases from 10{,}843 in SULAND\_v1 to 12{,}433 in SULAND\_v2, corresponding to a 14.7\% net increase. This increase should not be interpreted as simply appending 1{,}590 new boxes to the original annotations. The re-annotation process simultaneously added missing targets, removed invalid annotations, and revised retained boxes and class labels, as quantified in Table~\ref{tab:annotation_revision}.

The largest changes in foreground/background composition and annotation count occur in the IID splits. For example, the number of foreground images in the IID training split increases from 5{,}234 to 6{,}272, while the number of background-only images decreases from 17{,}522 to 16{,}484. Similar changes occur in the IID validation and test splits, indicating that a substantial number of samples previously treated as background-only contained visible targets. In contrast, the OOD split shows only modest changes in foreground coverage and total annotation count, suggesting that its refinement was driven primarily by label-consistency correction rather than large-scale recovery of missing targets.

Table~\ref{tab:annotation_revision} provides a finer box-level breakdown of the \emph{added}, \emph{removed}, \emph{tightened}, and \emph{class-corrected} scenarios defined in the preceding subsection. For matched annotations, \emph{Med.\ IoU} measures the median spatial overlap between the SULAND\_v1 and SULAND\_v2 boxes, while \emph{Med.\ ratio} denotes the median SULAND\_v2-to-SULAND\_v1 box-area ratio. The \%IoU$<$0.7 and \%shrunk columns report the proportions of matched boxes exhibiting substantial geometric change and reduced box area, respectively. The IID splits contain substantial numbers of both added and removed annotations, confirming that the refinement addressed missing targets as well as invalid positives. In the IID training split alone, 3{,}134 boxes were added, and 1{,}930 were removed.

The geometric statistics further show that many retained annotations required meaningful localization revision. Across the IID splits, the median IoU between matched SULAND\_v1 and SULAND\_v2 boxes is approximately 0.72--0.73, with 40.4--44.9\% of the matched boxes falling below an IoU of 0.7. The median area ratios range from 0.85 to 0.91, while 68.0--77.9\% of the revised boxes became smaller. These results indicate that the predominant geometric correction was the tightening of oversized or loosely positioned boxes around the visible target extent.

The OOD split exhibits a different correction pattern. Its higher median IoU of 0.82, median area ratio of 0.97, and lower proportion of boxes below 0.7 IoU (19.4\%) indicate that its bounding-box geometry was generally more consistent with the refined annotations than that of the IID splits. Nevertheless, 3{,}162 annotations required class correction. The high file-modification rate, therefore, results primarily from the systematic class-ID mismatch rather than extensive geometric revision.

As an additional measure of agreement, the SULAND\_v1 annotations were treated as predictions and evaluated against SULAND\_v2 using class-aware matching at IoU $\geq 0.5$. Across the combined IID splits, the original annotations achieved 62.1\% precision, 51.1\% recall, and an F1 score of 56.1\%. These results indicate that the differences between the two versions extend beyond isolated corrections and affect a substantial portion of the benchmark.

\begin{figure*}[!t]
  \centering
  \includegraphics[width=0.98\linewidth]{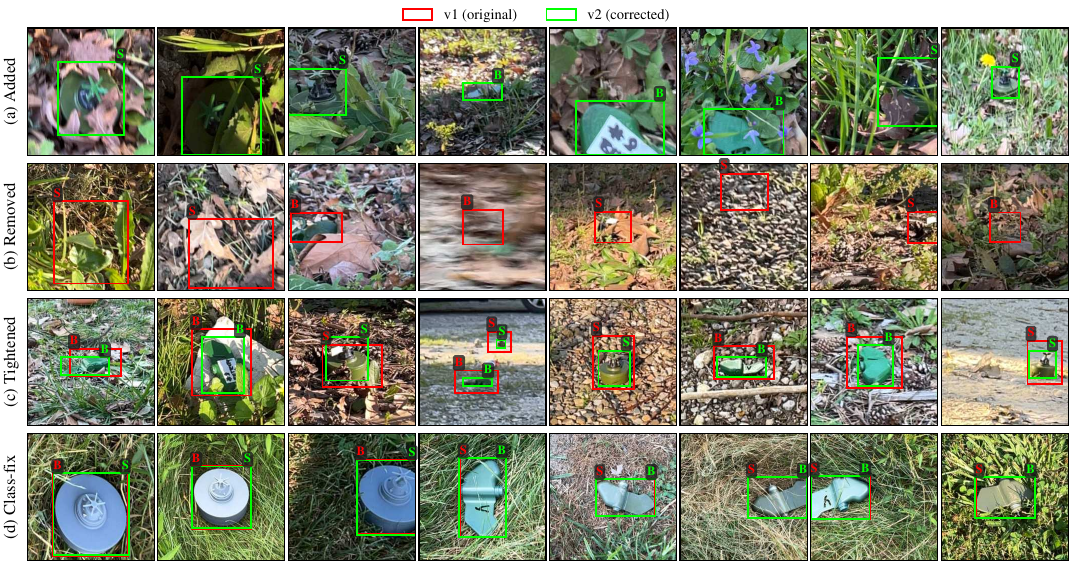}
  \caption{Representative annotation corrections from SULAND\_v1 to SULAND\_v2. Red and green boxes denote the original and refined annotations, respectively; B and S indicate PFM-1 (Butterfly) and PMA-2 (Starfish). Rows show added, removed, tightened, and class-corrected annotations.}

\label{fig:annotation_corrections}
\end{figure*}

The quantitative changes are further illustrated by the representative examples in Fig.~\ref{fig:annotation_corrections}. The original SULAND\_v1 annotations are shown in red and the corresponding SULAND\_v2 annotations in green, with B and S denoting PFM-1 (Butterfly) and PMA-2 (Starfish), respectively. The examples show how the refinement affected annotation completeness, bounding-box localization, and class consistency in representative samples.

Together, Tables~\ref{tab:v1v2_stats} and~\ref{tab:annotation_revision}, along with Fig.~\ref{fig:annotation_corrections}, show that SULAND\_v2 improves annotation completeness, removes invalid labels, refines bounding-box localization, and harmonizes the class convention across the IID and OOD splits.

\subsection{Impact of Annotation Refinement on Detector Performance}
\label{subsec:impact_dataset_cleaning}

  \begin{table*}[!t]
    \centering
    \caption{Impact of annotation refinement on YOLOv8 performance across five model scales. Each configuration was trained and evaluated using the same SULAND version. Values report mAP@50 (\%), and $\Delta$ denotes the absolute percentage-point change from SULAND\_v1 to SULAND\_v2.}
    \label{tab:v1_v2_comparison}
    \resizebox{\textwidth}{!}{%
    \begin{tabular}{ll ccc ccc ccc}
    \toprule
    & &
    \multicolumn{3}{c}{\textbf{mAP@50 (IID Val)}} &
    \multicolumn{3}{c}{\textbf{mAP@50 (IID Test)}} &
    \multicolumn{3}{c}{\textbf{mAP@50 (OOD Val)}} \\
    \cmidrule(lr){3-5}\cmidrule(lr){6-8}\cmidrule(lr){9-11}
    \textbf{Method} & \textbf{Backbone} &
    SULAND\_v1 & SULAND\_v2 & $\Delta$ &
    SULAND\_v1 & SULAND\_v2 & $\Delta$ &
    SULAND\_v1 & SULAND\_v2 & $\Delta$ \\
    \midrule
    \multirow{5}{*}{YOLOv8~\cite{yolov8}}
     & Nano   (N) & 66.9 & 77.2 & +10.3 & 70.1 & 84.7 & +14.6 & 12.9 & 30.7 & +17.8 \\
     & Small  (S) & 66.5 & 82.2 & +15.7 & 70.1 & 87.9 & +17.8 & 11.5 & 39.9 & +28.4 \\
     & Medium (M) & 68.0 & 82.2 & +14.2 & 69.2 & 88.8 & +19.6 &  9.8 & 48.6 & +38.8 \\
     & Large  (L) & 67.5 & 82.8 & +15.3 & 68.4 & 87.9 & +19.5 & 12.0 & 47.5 & +35.5 \\
     & XLarge (X) & 66.2 & 82.6 & +16.4 & 69.2 & 88.5 & +19.3 &  4.4 & 48.0 & +43.6 \\
    \bottomrule
    \end{tabular}}
  \end{table*}

To maintain comparability with the original SULAND\_v1 study, which evaluated the YOLOv8 Nano and Small configurations, we use the same detector family and extend the analysis to the Medium, Large, and XLarge model scales. Each configuration was trained and evaluated separately on SULAND\_v1 and SULAND\_v2 using the experimental settings described in Section~\ref{subsec:detector_protocol}. Table~\ref{tab:v1_v2_comparison} reports the resulting mAP@50 values and the corresponding absolute changes between the two dataset versions.

Annotation refinement improves performance consistently across all model scales and evaluation splits. On the IID validation and test sets, the gains range from 10.3--16.4 and 14.6--19.6 percentage points, respectively. This consistent improvement indicates that the effect of annotation refinement is not limited to a particular model capacity, with the Medium and Large configurations achieving the strongest IID performance on SULAND\_v2.

The most pronounced changes occur on the OOD split, where the improvements range from 17.8 to 43.6 percentage points. SULAND\_v1 yields uniformly low OOD scores across the five configurations, whereas SULAND\_v2 produces substantially higher values, reaching 48.6\% mAP@50 for YOLOv8-M. These differences should not be interpreted solely as improved model generalization, since the refined OOD annotations also correct the systematic class-ID mismatch identified in SULAND\_v1. Nevertheless, a substantial gap between IID and OOD performance remains after refinement, indicating that annotation correction improves benchmark reliability without eliminating the underlying domain shift.

Overall, Table~\ref{tab:v1_v2_comparison} shows that annotation quality materially affects both reported detector performance and the interpretation of OOD robustness. SULAND\_v2 therefore provides a more consistent basis for comparing detector configurations under IID and OOD evaluation.

\subsection{Cross-Version Training and Evaluation of SULAND\_v1 and SULAND\_v2}
\label{subsec:v1_v2_cross_evaluation_protocol}

  \begin{table*}[!t]
    \centering
    \caption{Cross-version YOLOv8 training and evaluation on SULAND\_v1 and SULAND\_v2 across five model scales. Values report mAP@50 (\%). The notation v1 $\!\to\! $v2 denotes training on SULAND\_v1 and evaluation using SULAND\_v2 annotations, with the remaining combinations defined analogously. The v1$^{\ast}$ results re-grade the same OOD predictions after correcting the inverted SULAND\_v1 class-ID convention.}
    \label{tab:cross_ablation_full}
    \resizebox{\textwidth}{!}{%
    \begin{tabular}{l cccc cccc cccccc}
    \toprule
    & \multicolumn{4}{c}{\textbf{mAP@50 (IID Val)}}
    & \multicolumn{4}{c}{\textbf{mAP@50 (IID Test)}}
    & \multicolumn{6}{c}{\textbf{mAP@50 (OOD Val)}} \\
    \cmidrule(lr){2-5}\cmidrule(lr){6-9}\cmidrule(lr){10-15}
    \textbf{Backbone}
    & v1$\to$v1 & v1$\to$v2 & v2$\to$v1 & v2$\to$v2
    & v1$\to$v1 & v1$\to$v2 & v2$\to$v1 & v2$\to$v2
    & v1$\to$v1 & v1$\to$v1$^{\ast}$ & v1$\to$v2 & v2$\to$v1 & v2$\to$v1$^{\ast}$ & v2$\to$v2 \\
    \midrule
    Nano   (N) & 66.9 & 50.5 & 39.0 & 77.2 & 70.1 & 60.1 & 40.3 & 84.7 & 12.9 & 33.2 & 36.6 &  8.7 & 26.7 & 30.7 \\
    Small  (S) & 66.5 & 51.0 & 42.7 & 82.2 & 70.1 & 60.4 & 42.3 & 87.9 & 11.5 & 35.0 & 39.0 &  8.9 & 35.9 & 39.9 \\
    Medium (M) & 68.0 & 54.5 & 42.2 & 82.2 & 69.2 & 61.7 & 41.9 & 88.8 &  9.8 & 31.2 & 33.7 & 11.0 & 43.3 & 48.6 \\
    Large  (L) & 67.5 & 51.6 & 45.1 & 82.8 & 68.4 & 59.3 & 45.6 & 87.9 & 12.0 & 41.2 & 45.1 &  9.0 & 42.5 & 47.5 \\
    XLarge (X) & 66.2 & 50.6 & 41.7 & 82.6 & 69.2 & 60.7 & 45.0 & 88.5 &  4.4 & 34.4 & 37.8 & 14.7 & 42.6 & 48.0 \\
    \midrule
    \textbf{Mean} & 67.0 & 51.6 & 42.1 & 81.4 & 69.4 & 60.4 & 43.0 & 87.5 & 10.1 & 35.0 & 38.4 & 10.5 & 38.2 & 42.9 \\
    \bottomrule
    \end{tabular}}

    \vspace{2pt}
  \end{table*}

The same-version comparison in Section~\ref{subsec:impact_dataset_cleaning} reflects the combined influence of training annotations and evaluation ground truth because each detector is trained and evaluated on the same dataset version. To separate these effects, we conduct a cross-version evaluation in which YOLOv8 models are trained independently on SULAND\_v1 and SULAND\_v2 under identical experimental settings and subsequently evaluated against both annotation versions without retraining.

Table~\ref{tab:cross_ablation_full} reports the four training--evaluation combinations for each model scale: v1$\to$v1, v1$\to$v2, v2$\to$v1, and v2$\to$v2. Here, v1$\to$v2 denotes training on SULAND\_v1 and evaluation using SULAND\_v2 annotations, with the remaining combinations interpreted analogously. Comparisons with a fixed evaluation version indicate the effect of the training annotations, whereas comparisons with a fixed trained model show how the evaluation annotations alone affect the reported score. Because both dataset versions contain the same images, changing only the evaluation version re-grades the same detections against different ground-truth annotations.

When evaluation is fixed to SULAND\_v2, training with the refined annotations consistently improves IID performance. The mean mAP@50 increases from 51.6\% for v1$\to$v2 to 81.4\% for v2$\to$v2 on IID validation and from 60.4\% to 87.5\% on IID testing. These differences indicate that the revised training annotations provide \textit{substantially more consistent supervision for in-distribution detection}. On the OOD split, the corresponding improvement is smaller, increasing from 38.4\% to 42.9\%, which suggests that annotation refinement improves OOD performance but does not remove the underlying cross-domain difficulty.

Holding the trained detector fixed also reveals strong sensitivity to the evaluation annotations. Models trained on SULAND\_v2 achieve mean IID validation and test scores of 81.4\% and 87.5\%, respectively, when evaluated against SULAND\_v2, but only 42.1\% and 43.0\% when the same detections are scored against SULAND\_v1. The lower cross-version scores should not be interpreted as poorer detector capability; rather, they show that differences in annotation completeness, localization, and consistency can substantially alter the measured performance.

The OOD results further isolate the effect of the inverted class-ID convention in SULAND\_v1. Using the original OOD labels, the mean scores are only 10.1\% for v1$\to$v1 and 10.5\% for v2$\to$v1. Re-grading the same predictions after correcting the class-ID mapping raises these values to 35.0\% (v1$\to$v1*) and 38.2\% (v2$\to$v1*), respectively. These corrected scores approach, but remain below, the corresponding evaluations against SULAND\_v2, indicating that the class-ID inversion accounts for most of the original OOD degradation, while the remaining difference reflects other annotation revisions.

Overall, the cross-evaluation results show that both training-label quality and evaluation-ground-truth consistency materially influence the reported benchmark performance. The consistently stronger v2$\to$v2 results, together with the large changes obtained by re-grading identical predictions, support the use of SULAND\_v2 as the common annotation standard for subsequent detector comparisons.

\section{Benchmarking Object Detectors Under IID and OOD Conditions}
\label{sec:benchmark_design_experimental_setup}
\subsection{Detector Families and Model Configurations}
\label{subsec:detector_families}
Although the original SULAND\_v1 study~\cite{Vivoli2024DeepImaging} evaluated only the Nano and Small variants of YOLOv8, the performance of broader detector families for RGB-based surface landmine detection, particularly under OOD conditions, has not been systematically established. We therefore evaluate 35 model configurations drawn from nine detector families, covering one-stage, two-stage, transformer-based, and open-vocabulary detection paradigms. This broader comparison examines whether findings obtained from a limited set of YOLOv8 configurations extend across architectures with different capacities, detection mechanisms, and computational characteristics.

The one-stage group includes YOLOv8~\cite{yolov8}, YOLO11~\cite{khanam2024yolov11}, YOLOv12~\cite{tian2026yolov12}, and YOLO26~\cite{sapkota2025yolo26}, each evaluated at the Nano, Small, Medium, Large, and XLarge scales. These configurations span a broad range of parameter counts and inference speeds, supporting comparison of detection performance and computational efficiency. YOLO-Worldv2~\cite{cheng2024yoloworld} is additionally evaluated at the Small, Medium, Large, and XLarge scales. Although YOLO-World incorporates vision--language representations and supports open-vocabulary detection, it is evaluated here under the same closed-set, two-class setting as the other methods to ensure comparability.

The two-stage group is represented by Faster R-CNN~\cite{2015_FasterRCNN} with ResNet-50 and ResNet-101 backbones. Its region-proposal-based formulation provides a useful contrast to the single-stage YOLO families, particularly for small and visually subtle targets, although it generally incurs greater computational cost.

The transformer-based group includes RT-DETR~\cite{zhao2024detrs} at the Large and XLarge scales, D-FINE~\cite{peng2024dfine} at the Nano, Small, Medium, Large, and XLarge scales, and RF-DETR~\cite{robinson2025rf} in Base and Large configurations using DINOv2-B and DINOv2-L backbones, respectively. These detectors formulate object detection through end-to-end set prediction and enable comparison across real-time and accuracy-oriented transformer designs.

All configurations are trained and evaluated on SULAND\_v2 using the common protocol described in the following subsection, Section~\ref{subsec:detector_protocol}. This standardized setup supports consistent assessment of IID and OOD detection performance, class-wise behavior, model size, and inference speed across substantially different detector architectures and capacities.

\subsection{Training and Evaluation Protocol}
\label{subsec:detector_protocol}

All 35 configurations spanning nine detector families and multiple model scales are fine-tuned from their publicly released pretrained weights on the SULAND\_v2 IID training split. Checkpoint selection is performed on the IID validation split using mAP@50-95, and final evaluation is conducted on the held-out IID test split and the OOD validation split. The original two-class task is retained throughout, with PFM-1 and PMA-2 as the only target categories. With the single exception of YOLO-Worldv2, every detector is initialized from COCO-pretrained~\cite{lin2014coco} weights before fine-tuning: the YOLO families (YOLOv8, YOLO11, YOLOv12, YOLO26), RT-DETR, and D-FINE use their official COCO detection checkpoints, Faster R-CNN uses the COCO-pretrained ResNet-50 model (the ResNet-101 variant instead uses an ImageNet-pretrained~\cite{5206848} backbone with a randomly initialized detection head), and RF-DETR pairs a COCO-pretrained detection head with a self-supervised DINOv2 backbone. YOLO-Worldv2 is instead initialized from its open-vocabulary pretraining on Objects365, GoldG, and CC3M-Lite, but is fine-tuned and evaluated here under the same closed-set two-class setting as the other detectors. Because all detectors begin from COCO-scale pretraining, the comparison isolates the effect of fine-tuning on SULAND\_v2 rather than differences in pretraining scale.

All models are trained under a common experimental setting to make the comparison as consistent as possible across frameworks. Models are trained for 100 epochs (50 for Faster R-CNN). This training budget is sufficient for every detector family to converge: each configuration reaches its peak validation mAP@50:95 well before the end of training and then plateaus, and we retain each model's best-epoch checkpoint rather than its final weights. The per-configuration convergence curves and this schedule-adequacy analysis are provided in the supplementary material (Training Convergence and Schedule Adequacy). The batch size is set to 16 and reduced to 8 for D-FINE-XLarge and RF-DETR-Large because of memory requirements. A fixed random seed of 42 is used for all experiments, and no augmentation is applied during evaluation.

Each detector family is trained using its standard optimizer and schedule. The Ultralytics-based detectors (the YOLO families, YOLO-Worldv2, and RT-DETR) are trained at an input size of $640{\times}640$ with the Ultralytics framework's automatically selected MuSGD optimizer (a Muon--SGD hybrid, momentum $0.9$) at a base learning rate of $0.01$. Faster R-CNN is trained with SGD using a learning rate of $0.005$, whereas D-FINE and RF-DETR are trained with AdamW; D-FINE follows the official per-size learning rates ($8\times10^{-4}$ for Nano, decreasing to $2.5\times10^{-4}$ for Large and XLarge) and RF-DETR uses $1\times10^{-4}$. Input resolutions follow each family's default configuration: $640{\times}640$ for the Ultralytics detectors (YOLO families, YOLO-Worldv2, and RT-DETR) and for D-FINE, $560{\times}560$ for RF-DETR (Base and Large), and a shorter-side resize to $800$ pixels with the longer side capped at $1333$ pixels for Faster R-CNN. All experiments are performed on a single NVIDIA A100 GPU.

\subsection{Evaluation Metrics}
\label{subsec:evaluation_metrics}

Detection performance is evaluated using mAP@50 and mAP@50:95. The former reports mean average precision at an intersection-over-union (IoU) threshold of 0.50, whereas the latter averages AP over IoU thresholds from 0.50 to 0.95 in increments of 0.05. Both metrics are included because mAP@50 facilitates comparison with prior landmine-detection studies, while mAP@50:95 provides a stricter assessment of bounding-box localization.

Precision and recall are reported at both the overall and class-specific levels. Precision measures the proportion of predicted detections that correspond to valid targets, whereas recall measures the proportion of ground-truth targets that are detected. Recall is particularly relevant to mine-action screening because missed targets may carry substantial operational consequences. For each detector, precision and recall are computed at the confidence threshold that maximizes the F1 score in IID validation split, thereby avoiding comparison at an arbitrary fixed threshold.

Model efficiency is characterized using the number of trainable parameters and inference throughput in frames per second (FPS). Together with the detection metrics, these quantities support evaluation of the tradeoff between predictive performance and computational cost. All accuracy metrics are reported separately for the IID and OOD evaluation subsets to assess changes in detection accuracy, localization quality, and target recovery under domain shift.

To ensure consistency across detector frameworks, all predictions are evaluated using a common Ultralytics-based evaluation routine~\cite{yolov8} rather than the native evaluator of each implementation. Predictions from non-Ultralytics models, including Faster R-CNN, D-FINE, and RF-DETR, are converted to a common format and scored using the same matching rules and metric computation. This unified procedure ensures that the reported results are directly comparable across detector families.

\section{Object Detectors Benchmark Results}
\label{sec:detector_results}

This section presents the performance of the evaluated detectors on SULAND\_v2. The overall benchmark first compares detection performance across detector families and model configurations. Subsequent subsections examine the IID--OOD generalization gap, the tradeoff between detection accuracy and inference speed, and class-wise performance for the PFM-1 and PMA-2 targets.

\subsection{Overall Detector Performance}
\label{subsec:overall_detector_performance}

  \begin{table*}[!t]
    \centering
    \caption{Detection performance of all evaluated methods and backbones on the SULAND dataset. Each metric shows IID (ITA $\to$ ITA in-distribution test)
  and OOD (ITA $\to$ USA out-of-distribution validation) results side by side. All models trained on SULAND\_v2. Best result per column shown in \textbf{bold}.}
    \label{tab:full_results}
    \begin{tabular}{ll cc cc cc cc cc}
    \toprule
    & &
    \multicolumn{2}{c}{\textbf{mAP@50}} &
    \multicolumn{2}{c}{\textbf{mAP@50-95}} &
    \multicolumn{2}{c}{\textbf{Recall}} &
    \multicolumn{2}{c}{\textbf{Precision}} & & \\
    \cmidrule(lr){3-4}\cmidrule(lr){5-6}\cmidrule(lr){7-8}\cmidrule(lr){9-10}
    \textbf{Method} & \textbf{Backbone} &
    IID & OOD & IID & OOD & IID & OOD & IID & OOD &
    \textbf{Params (M)} & \textbf{FPS} \\
    \midrule
    \multirow{5}{*}{YOLOv8~\cite{yolov8}}
     & Nano   (N) & 0.847 & 0.307 & 0.604 & 0.203 & 0.793 & 0.261 & 0.899 & 0.537 & 3.01 & 621.8 \\
     & Small  (S) & 0.879 & 0.399 & 0.651 & 0.273 & 0.820 & 0.270 & 0.913 & 0.707 & 11.13 & 532.3 \\
     & Medium (M) & 0.888 & 0.486 & 0.664 & 0.357 & 0.830 & 0.316 & 0.909 & 0.706 & 25.84 & 357.4 \\
     & Large  (L) & 0.879 & 0.475 & 0.669 & 0.350 & 0.821 & 0.379 & 0.907 & 0.445 & 43.61 & 264.8 \\
     & XLarge (X) & 0.885 & 0.480 & 0.665 & 0.339 & 0.827 & 0.416 & 0.903 & 0.441 & 68.13 & 185.0 \\
     \midrule
    \multirow{5}{*}{YOLO11~\cite{khanam2024yolov11}}
     & Nano   (N) & 0.885 & 0.419 & 0.659 & 0.295 & 0.829 & 0.283 & 0.911 & 0.660 & 2.58 & 567.9 \\
     & Small  (S) & 0.881 & 0.494 & 0.653 & 0.357 & 0.818 & 0.419 & 0.903 & 0.490 & 9.41 & 506.6 \\
     & Medium (M) & 0.863 & 0.391 & 0.641 & 0.261 & 0.814 & 0.280 & 0.876 & 0.642 & 20.03 & 342.6 \\
     & Large  (L) & 0.906 & 0.529 & 0.694 & 0.389 & 0.843 & 0.430 & 0.900 & 0.537 & 25.28 & 294.1 \\
     & XLarge (X) & 0.899 & 0.438 & 0.678 & 0.317 & 0.829 & 0.402 & 0.910 & 0.408 & 56.83 & 183.8 \\
     \midrule
    \multirow{5}{*}{YOLOv12~\cite{tian2026yolov12}}
     & Nano   (N) & 0.878 & 0.386 & 0.645 & 0.271 & 0.834 & 0.362 & 0.875 & 0.324 & 2.56 & 511.9 \\
     & Small  (S) & \textbf{0.908} & 0.470 & 0.687 & 0.338 & \textbf{0.847} & 0.397 & 0.918 & 0.459 & 9.23 & 401.1 \\
     & Medium (M) & 0.889 & 0.519 & 0.663 & 0.358 & 0.829 & 0.347 & 0.904 & 0.744 & 20.11 & 257.8 \\
     & Large  (L) & 0.891 & 0.456 & 0.675 & 0.326 & 0.826 & 0.306 & 0.916 & 0.673 & 26.34 & 178.2 \\
     & XLarge (X) & 0.893 & 0.423 & 0.675 & 0.304 & 0.825 & 0.361 & 0.910 & 0.390 & 59.05 & 117.1 \\
     \midrule
    \multirow{5}{*}{YOLO26~\cite{sapkota2025yolo26}}
     & Nano   (N) & 0.874 & 0.235 & 0.617 & 0.143 & 0.785 & 0.194 & 0.912 & 0.495 & 2.38 & \textbf{639.2} \\
     & Small  (S) & 0.888 & 0.315 & 0.683 & 0.219 & 0.811 & 0.188 & 0.900 & 0.591 & 9.47 & 524.0 \\
     & Medium (M) & 0.856 & 0.360 & 0.638 & 0.260 & 0.784 & 0.272 & 0.885 & 0.340 & 20.35 & 360.4 \\
     & Large  (L) & 0.901 & 0.437 & 0.693 & 0.303 & 0.836 & 0.367 & 0.900 & 0.410 & 24.75 & 298.1 \\
     & XLarge (X) & 0.902 & 0.477 & \textbf{0.708} & 0.357 & 0.816 & 0.316 & 0.916 & 0.555 & 55.64 & 190.4 \\
     \midrule
    \multirow{4}{*}{YOLO-Worldv2~\cite{cheng2024yoloworld}}
     & Small  (S) & 0.884 & 0.417 & 0.653 & 0.298 & 0.844 & 0.273 & 0.903 & 0.679 & 12.75 & 458.4 \\
     & Medium (M) & 0.895 & 0.486 & 0.672 & 0.351 & 0.833 & 0.342 & 0.922 & 0.533 & 28.36 & 326.1 \\
     & Large  (L) & 0.887 & 0.468 & 0.668 & 0.332 & 0.818 & 0.343 & 0.925 & 0.474 & 46.81 & 235.1 \\
     & XLarge (X) & 0.893 & 0.513 & 0.678 & 0.379 & 0.793 & 0.386 & \textbf{0.951} & 0.520 & 72.86 & 169.7 \\
     \midrule
    \multirow{2}{*}{RT-DETR~\cite{zhao2024detrs}}
     & Large  & 0.809 & 0.222 & 0.550 & 0.143 & 0.749 & 0.215 & 0.928 & 0.614 & 31.99 & 150.7 \\
     & XLarge & 0.788 & 0.187 & 0.534 & 0.135 & 0.702 & 0.184 & 0.928 & 0.540 & 65.47 & 109.5 \\
     \midrule
    \multirow{2}{*}{Faster R-CNN~\cite{2015_FasterRCNN}}
     & ResNet-50  & 0.851 & 0.723 & 0.626 & 0.560 & 0.750 & 0.599 & 0.925 & 0.785 & 43.26 & 53.2 \\
     & ResNet-101 & 0.865 & 0.746 & 0.598 & 0.510 & 0.785 & 0.575 & 0.901 & 0.859 & 60.24 & 51.7 \\
     \midrule
    \multirow{5}{*}{D-FINE~\cite{peng2024dfine}}
     & Nano   (N) & 0.788 & 0.295 & 0.567 & 0.209 & 0.756 & 0.267 & 0.917 & 0.558 & 3.72 & 444.8 \\
     & Small  (S) & 0.757 & 0.433 & 0.551 & 0.344 & 0.726 & 0.350 & 0.902 & 0.814 & 10.21 & 366.1 \\
     & Medium (M) & 0.755 & 0.464 & 0.554 & 0.378 & 0.731 & 0.391 & 0.921 & 0.911 & 19.44 & 243.1 \\
     & Large  (L) & 0.777 & 0.569 & 0.578 & 0.467 & 0.726 & 0.537 & 0.939 & 0.941 & 31.08 & 170.5 \\
     & XLarge (X) & 0.779 & 0.449 & 0.568 & 0.358 & 0.764 & 0.361 & 0.911 & \textbf{0.978} & 62.46 & 108.6 \\
     \midrule
    \multirow{2}{*}{RF-DETR~\cite{robinson2025rf}}
     & Base  (DINOv2-B) & 0.848 & 0.600 & 0.591 & 0.458 & 0.794 & 0.468 & 0.922 & 0.746 & 31.85 & 25.9 \\
     & Large (DINOv2-L) & 0.880 & \textbf{0.799} & 0.631 & \textbf{0.623} & 0.805 & \textbf{0.675} & 0.932 & 0.883 & 135.16 & 22.2 \\
    \bottomrule
    \end{tabular}
  \end{table*}

Table~\ref{tab:full_results} summarizes the performance of all evaluated configurations on the IID test and OOD validation splits. Most detector families achieve strong IID performance, indicating that the surface-mine detection task can be learned effectively when the evaluation conditions remain similar to those represented during training. The strongest IID results are obtained by the YOLO families: YOLOv12-S achieves the highest mAP@50 of 0.908, whereas YOLO26-X achieves the highest mAP@50:95 of 0.708. Faster R-CNN and RF-DETR also remain competitive under IID evaluation, although at substantially lower inference speeds.

The detector ranking changes under OOD evaluation. RF-DETR-L achieves the strongest overall OOD performance, leading in mAP@50, mAP@50:95, and recall. Faster R-CNN also performs consistently across both backbones, while D-FINE-L provides the strongest OOD result among the remaining transformer configurations. In comparison, most YOLO variants remain below approximately 0.53 OOD mAP@50 despite their strong IID accuracy. These results show that the architectures that perform best under IID conditions are not necessarily those that retain the highest accuracy under distribution shift.

The precision and recall results further show that a high value for one metric does not necessarily imply balanced detection behavior. For example, some configurations maintain high OOD precision but recover a comparatively smaller proportion of the targets. Detector comparison should therefore consider mAP, precision, and recall jointly rather than relying on a single metric. The observed changes in model ranking and metric balance are examined further through the IID--OOD generalization analysis.

\subsection{IID--OOD Generalization Gap}
\label{subsec:iid_ood_generalization}

\begin{figure*}[!t]
  \centering
  \begin{subfigure}{0.49\linewidth}
    \centering
    \includegraphics[width=\linewidth]{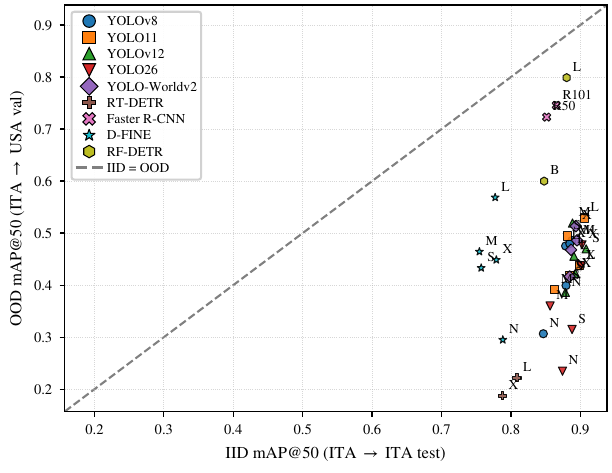}
    \label{fig:iid_vs_ood_map50}
  \end{subfigure}
  \hfill
  \begin{subfigure}{0.49\linewidth}
    \centering
    \includegraphics[width=\linewidth]{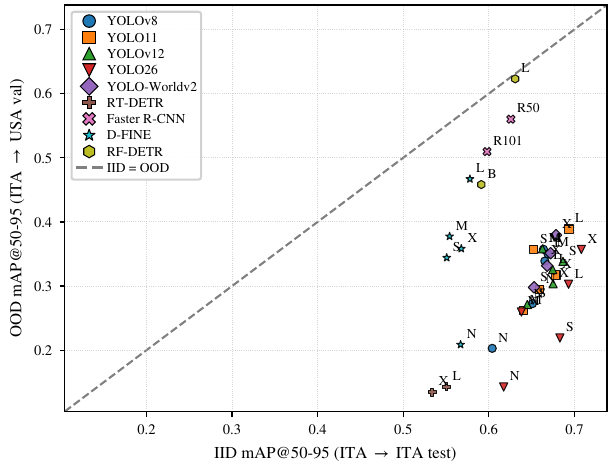}
    \label{fig:iid_vs_ood_map5095}
  \end{subfigure}
  \vspace{-0.1in}
  \caption{IID and OOD detection performance for all models trained on SULAND\_v2, shown using mAP@50 (left) and mAP@50:95 (right). Each point represents one model configuration, and marker shape denotes the detector family. The dashed diagonal indicates equal IID and OOD performance; greater displacement below the diagonal corresponds to a larger performance reduction under domain shift.}
  \label{fig:iid_vs_ood}
\end{figure*}

Table~\ref{tab:full_results} provides the numerical comparison between IID and OOD performance, while Fig.~\ref{fig:iid_vs_ood} visualizes the corresponding generalization gap. Each point represents one trained configuration, and the dashed diagonal indicates equal performance under the two evaluation conditions. All configurations fall below this line, showing that every evaluated detector experiences an accuracy reduction on the OOD split. However, the magnitude of this reduction varies considerably across detector families.

Most YOLO-family and RT-DETR configurations occupy the high-IID, lower-OOD region of the plots, indicating considerable sensitivity to the distribution shift. In contrast, RF-DETR-L and the Faster R-CNN variants remain substantially closer to the diagonal and preserve a larger proportion of their IID performance. RF-DETR-L exhibits the strongest retention, decreasing from 0.880 to 0.799 mAP@50, while its mAP@50:95 changes only from 0.631 to 0.623. D-FINE-L also retains a comparatively large proportion of its IID performance relative to most high-throughput one-stage configurations.

The mAP@50:95 results show that the effect of domain shift extends beyond target recognition at an IoU threshold of 0.50 to localization under stricter overlap requirements. Model scale alone does not explain the observed robustness, since larger variants do not consistently outperform smaller configurations within the same family. This pattern suggests that OOD generalization depends more strongly on detector architecture and learned representation than on parameter count alone. These findings reinforce the importance of reporting OOD performance separately rather than treating high IID accuracy as evidence of deployment robustness.

\subsection{Accuracy--Speed Tradeoff}
\label{subsec:accuracy_speed_tradeoff}

\begin{figure*}[!t]
  \centering
  \includegraphics[width=\linewidth]{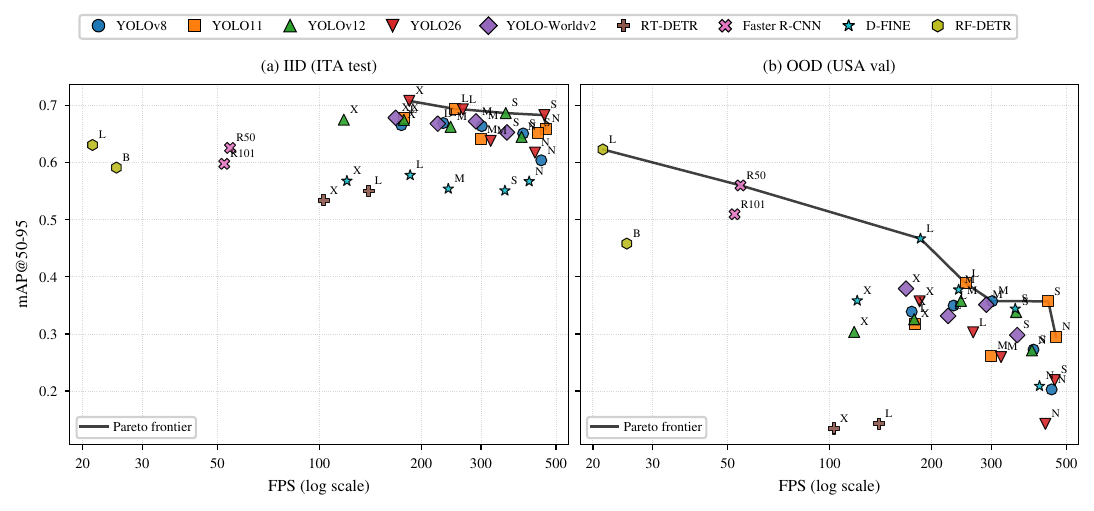}
  \caption{Accuracy--speed tradeoff for detectors trained on SULAND\_v2 under IID (left) and OOD (right) evaluation. Accuracy is measured using mAP@50:95, and inference speed is shown on a logarithmic FPS axis. The black curve connects the nondominated configurations on the Pareto frontier.}
  \label{fig:pareto}
\end{figure*}

Figure~\ref{fig:pareto} relates mAP@50:95 to inference speed under IID and OOD evaluation. The Pareto frontier contains the nondominated configurations for which no other evaluated detector is both faster and more accurate. Under IID evaluation, the frontier is dominated primarily by YOLO-family configurations, which combine high detection accuracy with inference rates of several hundred frames per second. Lightweight YOLO models are therefore attractive for high-throughput processing when deployment conditions remain similar to those represented during training.

The OOD setting produces a substantially different tradeoff. RF-DETR-L and Faster R-CNN occupy the high-accuracy, lower-throughput region, while D-FINE-L provides an intermediate compromise between OOD accuracy and inference speed. Higher-throughput YOLO configurations remain computationally attractive and occupy the faster portion of the tradeoff, although their OOD accuracy remains below that of RF-DETR-L and Faster R-CNN. Thus, the configurations providing the strongest IID accuracy--speed balance are not necessarily those that preserve performance most effectively under domain shift.

The parameter counts reported in Table~\ref{tab:full_results} further show that model size is not a consistent predictor of detection accuracy or OOD robustness. Increasing model scale within a detector family does not uniformly improve either IID or OOD performance. Detector selection should therefore account for the intended operating conditions: lightweight YOLO configurations may be appropriate for rapid processing in familiar environments, whereas slower but more robust architectures may be preferable when performance must be preserved across previously unseen conditions.

\subsection{Class-Wise Performance}
\label{subsec:class_wise_performance}

 \begin{figure*}[!t]
  \centering
  \includegraphics[width=0.96\linewidth]{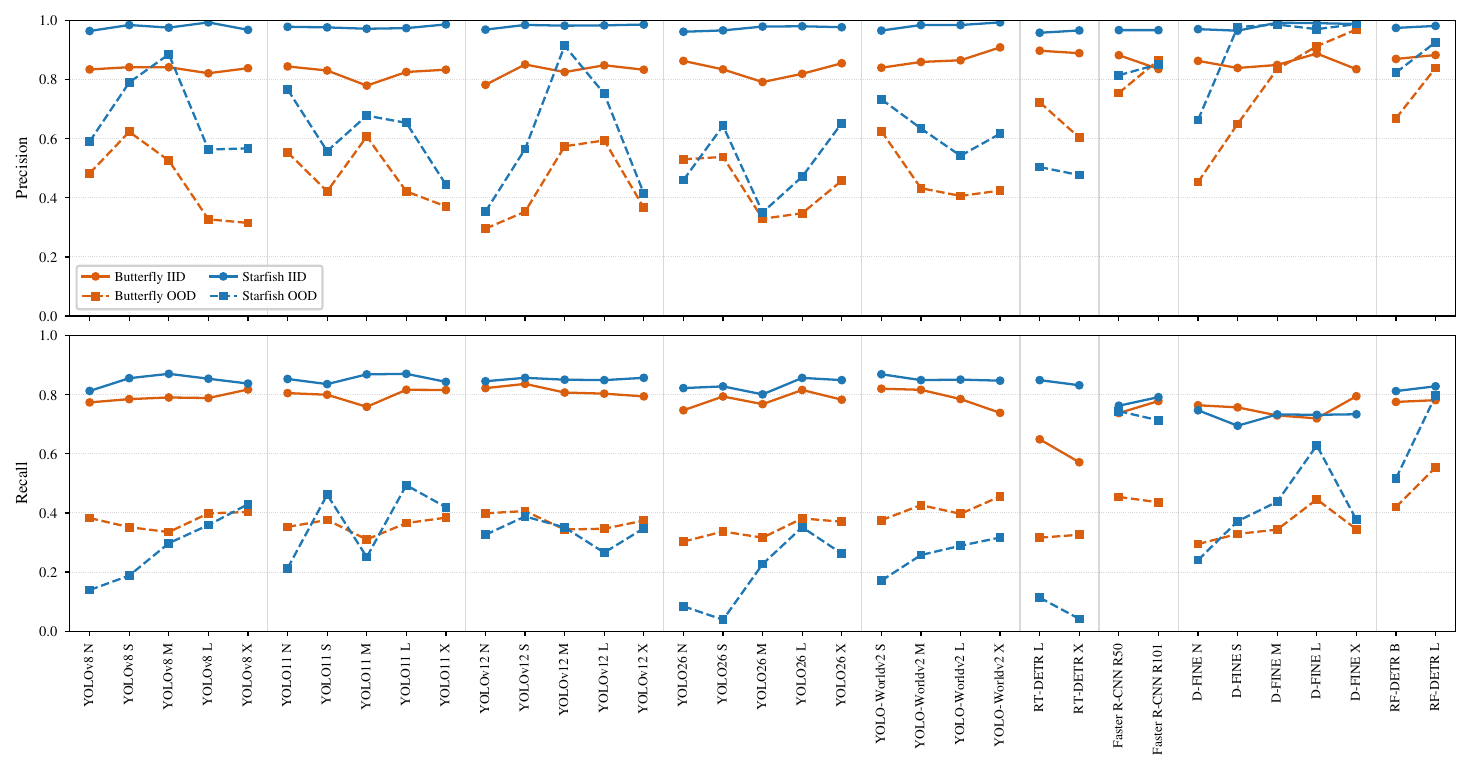}
  \caption{Per-class precision and recall for PFM-1 (\emph{Butterfly}) and PMA-2 (\emph{Starfish}) under IID and OOD evaluation. Solid lines denote IID results and dashed lines denote OOD results; colors indicate the target class.}
  \label{fig:per_class_pr}
\end{figure*}

Figure~\ref{fig:per_class_pr} presents class-specific precision and recall for PFM-1 (\emph{Butterfly}) and PMA-2 (\emph{Starfish}) under IID and OOD evaluation. Under IID conditions, most detector configurations achieve consistently high precision and recall for both target classes, indicating that both classes can be detected reliably when the evaluation conditions remain similar to those represented during training.

Under OOD evaluation, both classes exhibit substantial performance degradation, although the effect differs across detector families and metrics. For many configurations, precision is better preserved than recall, indicating that the detectors remain relatively selective when producing predictions but fail to recover a considerable proportion of the targets. Several YOLO configurations, for example, retain comparatively high PMA-2 precision while exhibiting much lower PMA-2 recall. This pattern suggests that the models produce fewer detections for this class under OOD conditions, but a relatively large proportion of those detections remain correct.

The class-wise behavior changes for the stronger OOD detectors. Faster R-CNN, D-FINE-L, and particularly RF-DETR-L retain higher recall than most YOLO and RT-DETR configurations, although the relative performance of PFM-1 and PMA-2 remains architecture-dependent. These results show that aggregate mAP alone does not fully characterize OOD behavior: detectors with similar overall performance may differ substantially in class-specific precision and target recovery. For mine-action screening, this distinction is important because \textit{low recall for either target class directly increases the risk of missed detections.}

\subsection{Qualitative Detection Results and Error Analysis}
\label{subsec:qualitative_error_analysis}

\begin{figure*}[!t]
  \centering
  \includegraphics[width=0.98\linewidth]{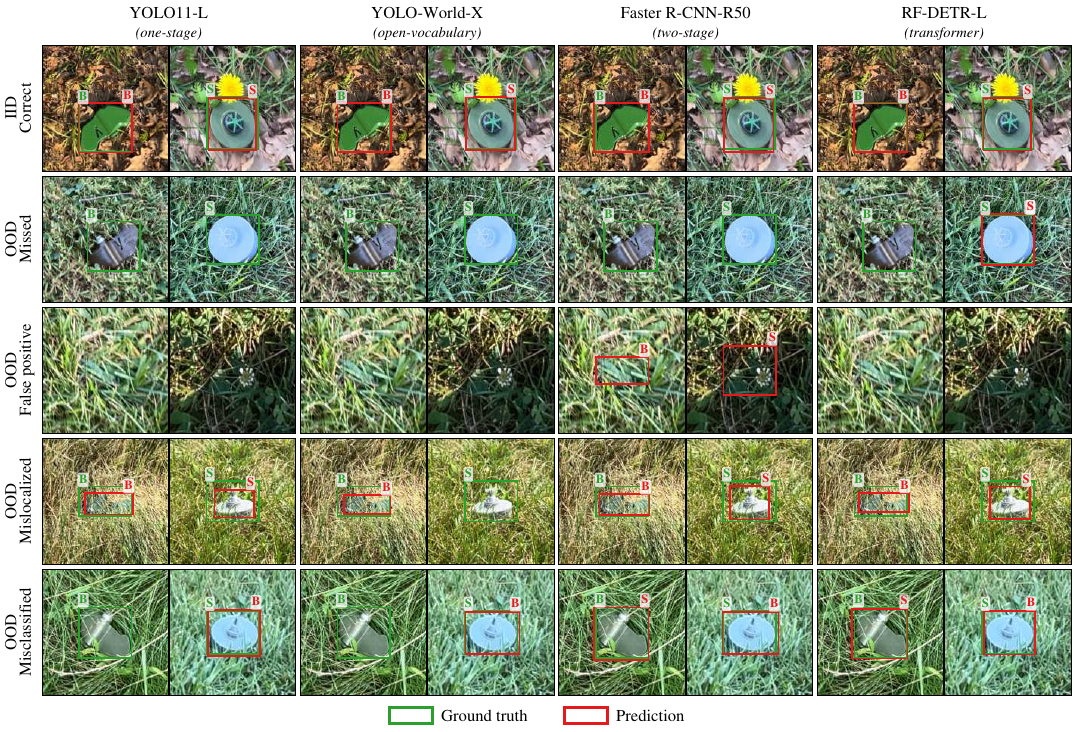}
  \caption{Representative IID and OOD detections for the best OOD configuration from each detector family, selected by mAP@50:95. Rows show case types and columns show detectors; green and red boxes denote ground truth and predictions. B and S denote PFM-1 (Butterfly) and PMA-2 (Starfish), respectively.}
  \label{fig:qualitative_detection_results}
\end{figure*}

To complement the quantitative benchmark, Fig.~\ref{fig:qualitative_detection_results} presents representative predictions from one configuration in each detector category. To maintain readability, the configuration with the highest OOD mAP@50:95 was selected from each category: YOLO11-L for one-stage detection, YOLO-Worldv2-X for open-vocabulary detection, Faster R-CNN-R50 for two-stage detection, and RF-DETR-L for transformer-based detection. The figure includes an IID reference case and representative OOD examples illustrating missed detections, false positives, localization errors, and class-assignment errors.

In the IID example, the selected detectors localize clearly visible targets under conditions similar to those represented during training. The OOD examples exhibit a broader range of prediction errors. Some targets are missed when their appearance is affected by background clutter, vegetation, shadows, reduced contrast, viewpoint variation, or partial occlusion. False positives occur in background regions containing target-like shapes or textures, while localization errors arise when predicted boxes do not adequately match the visible target extent. The illustrated cases also include incorrect class assignments between PFM-1 and PMA-2.

Differences among detector categories are also visible in the selected examples. RF-DETR-L and Faster R-CNN-R50 recover more of the challenging targets in several illustrated OOD scenes, whereas YOLO11-L and YOLO-Worldv2-X exhibit more missed detections or greater sensitivity to background variation in those cases. These observations are consistent with the aggregate OOD results, but the selected examples are not intended to establish that one detector will outperform another in every scene.

Taken together, these examples provide a qualitative counterpart to the quantitative benchmark by illustrating how the observed OOD performance differences manifest at the prediction level across representative detector categories. The selected cases show that detectors with different aggregate OOD performance can exhibit different combinations of missed detections, background-induced false positives, localization errors, and class-assignment errors. They are not intended to quantify the frequency or establish the causes of each failure mode, but to visually support and contextualize the detector-level and class-wise results reported in the preceding subsections.

\section{Discussion}
\label{sec:discussion}

The findings of this study indicate that benchmark reliability and robustness evaluation should be considered jointly in RGB-based surface-mine detection. Annotation inconsistencies can affect both the supervision used during training and the reference labels used during evaluation, while IID testing alone may not reveal how a detector behaves under changed environmental and acquisition conditions. SULAND\_v2 therefore provides a more consistent basis for examining detector performance across both IID and OOD settings.

\subsection{Benchmark Reliability and Dataset Curation}
\label{subsec:dataset_quality_benchmark_reliability}

The SULAND\_v1 audit shows that annotation quality is an important component of benchmark validity. Missing targets, invalid annotations, inconsistent localization criteria, and class-convention differences influence training and evaluation in different ways. Errors in the training labels alter the supervision provided to the detector, whereas errors in the evaluation labels can penalize valid predictions or reward incorrect ones. Consequently, measured performance may partly reflect compatibility with a particular annotation convention rather than target-detection capability alone.

The cross-version experiments further illustrate this distinction. Models trained and evaluated using the same annotation version may reproduce version-specific labeling patterns, whereas evaluation using a different annotation version exposes disagreement between the learned supervision and the reference labels. Performance changes across dataset versions should therefore not be interpreted solely as differences in model capability; they may also reflect changes in annotation completeness, localization criteria, and class consistency.

This observation is particularly relevant to remote-sensing applications in which only a limited number of public datasets are available. Undetected annotation inconsistencies can propagate across subsequent studies and influence reported model rankings. Benchmark releases should therefore document class definitions, annotation criteria, correction procedures, dataset versions, and evaluation protocols. By retaining the original imagery and split organization while applying a unified annotation procedure, SULAND\_v2 provides a more consistent basis for comparison without redefining the underlying detection task.

\subsection{Implications for Robustness Evaluation}
\label{subsec:discussion_ood_benchmarking}

The change in detector ranking between IID and OOD evaluation shows that performance on a single test distribution is insufficient for characterizing deployment robustness. Strong IID performance demonstrates that a model can learn the visual patterns represented in the training data, but it does not establish that the same behavior will be preserved under changes in location, terrain, background, illumination, viewpoint, or target appearance. OOD evaluation therefore provides information that cannot be inferred from IID accuracy alone.

The results also indicate that increasing model scale is not a consistent solution to domain shift. Larger configurations do not uniformly retain more accuracy than smaller variants from the same family. The observed differences among detector families may be associated with variations in feature representation, multiscale processing, proposal generation, and query-based detection. However, the present experiments do not isolate the contribution of these components, and controlled architectural ablations would be required to establish causal explanations.

These findings support the separate reporting of IID and OOD results rather than combining them into a single aggregate measure. Overall mAP should also be considered together with localization-sensitive metrics and class-specific precision and recall, because detectors with similar aggregate performance may exhibit different error patterns. Future benchmark protocols would benefit from multiple OOD domains so that robustness can be examined across distinct geographic, seasonal, environmental, and acquisition changes rather than through a single transfer setting.

\subsection{Operational Relevance for Mine Action}
\label{subsec:operational_implications_mine_action}

The operational relevance of a detector depends not only on average accuracy but also on the consequences of its errors. False alarms increase the effort required for follow-up inspection, whereas missed detections may leave potential hazards unflagged. The observed reduction in OOD recall is therefore important: a detector may remain selective when producing predictions while failing to recover a sufficient proportion of visible targets under shifted conditions.

The accuracy--speed results also show that computational efficiency and OOD robustness represent different design objectives. High-throughput configurations may be useful for rapid preliminary screening or repeated processing in familiar environments, whereas more computationally demanding models may be preferable when preserving performance in a new environment is the primary concern. A tiered workflow could use a fast detector for initial candidate generation, followed by a detector that demonstrated stronger OOD performance in this benchmark or by a human analyst for secondary review. This workflow is an operational implication of the results rather than a configuration directly evaluated in this study.

Operational deployment would also require threshold selection based on application risk rather than only the F1-optimal operating points used for benchmark comparison. Recall-oriented thresholds, confidence calibration, and the expected number of false alarms per surveyed area should be examined before field use. Appropriate operating points may differ across survey stages, target types, and environmental conditions.

These findings support the use of RGB detection as a decision-support component for prioritizing inspection, documenting visible surface conditions, and guiding closer examination. They do not support its use as an autonomous basis for declaring an area safe. RGB imagery remains limited to hazards that produce visible surface evidence and cannot address fully buried, heavily occluded, or visually indistinguishable targets without complementary sensing and established mine-action procedures.

\subsection{Limitations and Future Directions}
\label{subsec:limitations}

The scope of the findings is constrained by the composition of SULAND\_v2. The dataset contains two surface-laid target classes represented by surrogate or inert objects, and the principal OOD evaluation is based on transfer between the released Italian and USA subsets. Although these subsets contain meaningful visual variation, they do not represent the full diversity of operational minefields.

The results may not transfer directly to other mine or UXO types, damaged or weathered objects, dense target arrangements, partially buried targets, severe occlusion, seasonal vegetation changes, different camera systems, or substantially different altitudes and viewing geometries. The use of RGB imagery also restricts the benchmark to visible surface cues and excludes hazards for which no reliable visual evidence is available.

Although the annotation refinement followed predefined criteria and included multi-reviewer quality control, the process remained primarily manual. Decisions involving small, blurred, partially occluded, or weakly contrasted targets can require subjective judgment, and some residual omissions, localization inconsistencies, or class-assignment errors may remain. SULAND\_v2 should therefore be regarded as a more consistent annotation release rather than an error-free reference. Future versions may benefit from continued community review, versioned correction records, and quantitative assessment of inter-annotator agreement.

The factors contributing to the IID--OOD difference are partially entangled. Geographic transfer occurs together with changes in terrain, vegetation, background composition, illumination, viewpoint, and target presentation. The feature-space analysis and detector results provide evidence of visual differences between the IID and OOD subsets and show that model performance changes across these conditions, but they do not quantify the contribution of each individual factor. The t-SNE visualization should therefore be interpreted as qualitative evidence rather than as a direct measure of domain divergence.

All detector configurations were evaluated using a common training and scoring protocol to support comparability. However, a standardized protocol may not provide the individually optimal setting for every detector family. Inference speed also depends on the hardware, software implementation, input resolution, and measurement procedure used in this study and should therefore be interpreted comparatively rather than as a universal deployment rate.

Future work should extend the benchmark across additional geographic regions, seasons, altitudes, camera geometries, target types, and levels of visibility. Cross-dataset evaluation will be important for determining whether the observed detector rankings remain stable on independently collected imagery. Factorized test sets that vary one environmental or acquisition condition at a time would also help identify the sources of generalization failure.

Methodological extensions should examine domain generalization, domain adaptation, test-time adaptation, few-shot learning, confidence calibration, and uncertainty-aware detection~\cite{Lekhak2025UncertaintyDropout}. Synthetic data and simulation-to-real approaches may help expand rare target and background combinations, but their utility should be validated using independently acquired real imagery. Finally, combining RGB detection with complementary sensing modalities may extend the resulting system beyond visible surface evidence and support a more comprehensive mine-action survey and decision-support workflow.

\section{Conclusion}

\label{sec:conclusion}

This study introduced SULAND\_v2, a refined version of the SULAND RGB surface-mine dataset \cite{Vivoli2024DeepImaging}, and used it to establish a standardized IID and OOD object-detection benchmark. The refinement addressed missing and invalid annotations, inconsistent bounding-box localization, partial-visibility criteria, and class-ID conventions while preserving the original imagery and split organization. Cross-version experiments showed that these annotation differences can materially affect both model training and reported evaluation outcomes, emphasizing the need to verify benchmark quality before interpreting detector rankings.

Using SULAND\_v2, 35 configurations from nine detector families were evaluated under a common protocol. Most models achieved strong IID performance, whereas their OOD behavior varied substantially across architectures. YOLO-family models generally provided the strongest IID accuracy and highest inference throughput, while RF-DETR-Large and Faster R-CNN preserved considerably more performance under the geographic and environmental shift represented by the OOD split. Increasing model scale did not consistently improve generalization, and the accuracy--speed analysis showed that the fastest configurations were not necessarily the most robust.

Overall, the findings show that reliable RGB-based surface-mine detection requires attention to both annotation quality and evaluation under distribution shift. SULAND\_v2 provides a more consistent basis for comparing detectors and for studying robustness beyond the training environment. RGB detection should be viewed as a survey-support and decision-support capability for visible surface targets rather than as an independent clearance method. Future work should extend the benchmark to additional geographic regions, target types, seasons, acquisition conditions, and sensing modalities, while investigating domain-generalization, adaptation, calibration, and uncertainty-aware detection methods.

\section{Dataset and Code Availability}
\label{sec:dataset_code_availability}
The SULAND\_v1 and SULAND\_v2 datasets are publicly available at
\url{https://huggingface.co/datasets/SagarLekhak/SULAND_v2_RGB_Surface_Landmine_Dataset}. The repository includes the original SULAND\_v1 annotations, the SULAND\_v1
OOD annotations with corrected class IDs, the refined SULAND\_v2 annotations,
and detailed sample-level annotation-audit records. The code used for detectors
training, evaluation, and benchmarking is available at
\url{https://github.com/PrasannaPulakurthi/SULAND_v2}.

\small
\bibliographystyle{IEEEtranN}
\bibliography{references}
\begin{IEEEbiography}[{\includegraphics[width=1in,height=1.25in,clip,keepaspectratio]{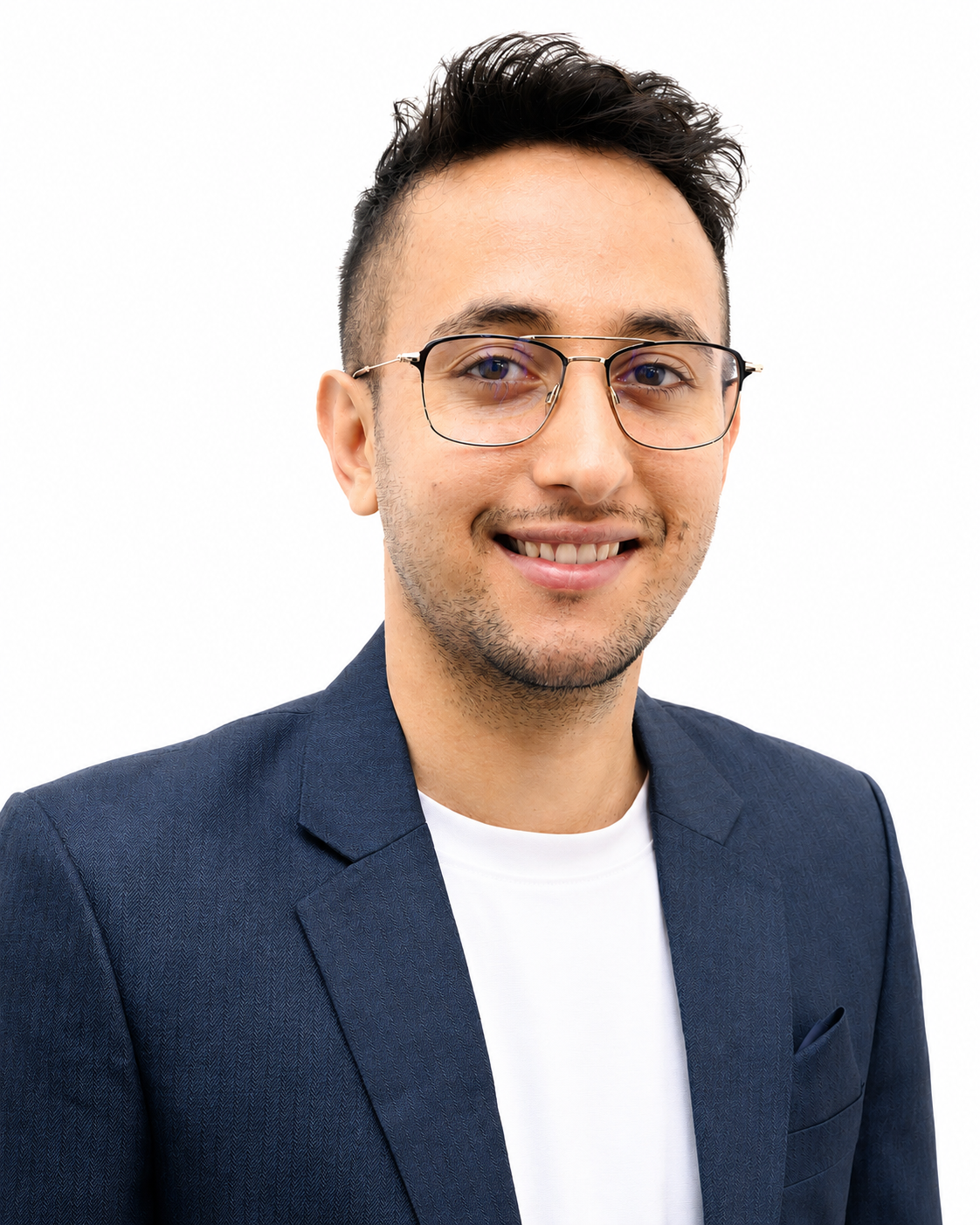}}]{Sagar Lekhak}
 received the B.E. degree in Electronics and Communication Engineering from Tribhuvan University, Nepal, in 2022, and the M.S. degree in Imaging Science from the Rochester Institute of Technology, Rochester, NY, USA, in 2025. He is currently pursuing the Ph.D. degree in Imaging Science at the Rochester Institute of Technology, where he has been a doctoral student since 2023.

His research interests include target detection, computer vision, deep learning, remote sensing, and hyperspectral imaging. His current research focuses on multimodal UAV-based landmine and unexploded ordnance detection using hyperspectral, multispectral, thermal, RGB, LiDAR, polarimetric, magnetometer, and metal detector data. His broader interests include imaging optics, deep learning for critical applications, and advanced sensing technologies for humanitarian demining and environmental monitoring.
\end{IEEEbiography}

\begin{IEEEbiography}[{\includegraphics[width=1in,height=1.25in, clip, keepaspectratio]{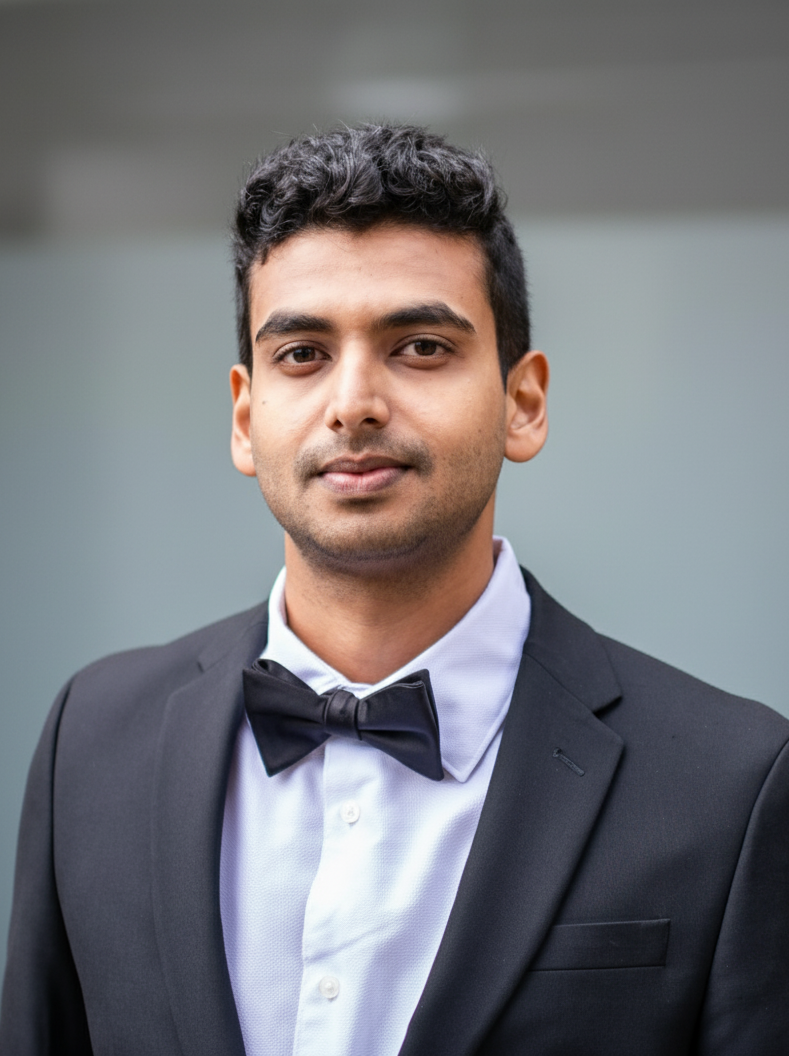}}]{Prasanna Reddy Pulakurthi} received the B.E. degree in electronics and communications engineering from PES University, Bengaluru, India, in 2017, and the M.S. degree in electrical and microelectronic engineering and the Ph.D. degree from the Rochester Institute of Technology, Rochester, NY, USA, in 2019 and 2025, respectively. His research interests include generative AI, computer vision, machine learning, and deep learning, with a focus on efficient generative modeling, source-free domain adaptation, human action recognition, and explainable multimodal (text-video) retrieval using large language models (LLMs).
\end{IEEEbiography}

\begin{IEEEbiography}[{\includegraphics[width=1in,height=1.25in,clip,keepaspectratio]{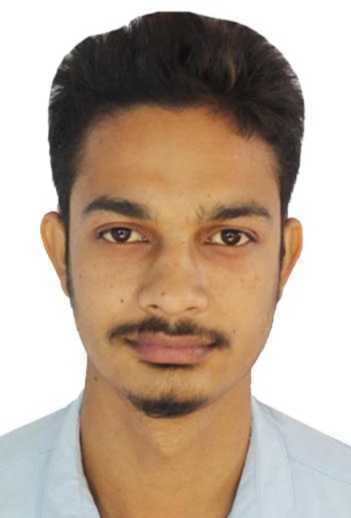}}]{Lalit Joshi}
  holds a B.E. in Electronics and Communication Engineering from Tribhuvan University, Nepal. He is currently pursuing the the Master of Science program in Informatics and Intelligent Systems Engineering in the Institute of Engineering, Thapathali Campus, Nepal. His areas of interest include Computer Vision, Generative Artificial Intelligence (GenAI), healthcare data analysis, and intelligent systems. Some other areas of his interests include AI driven healthcare applications, and the development of intelligent technologies for real world problem solving.
\end{IEEEbiography}

\begin{IEEEbiography}[{\includegraphics[width=1in,height=1.25in,clip,keepaspectratio]{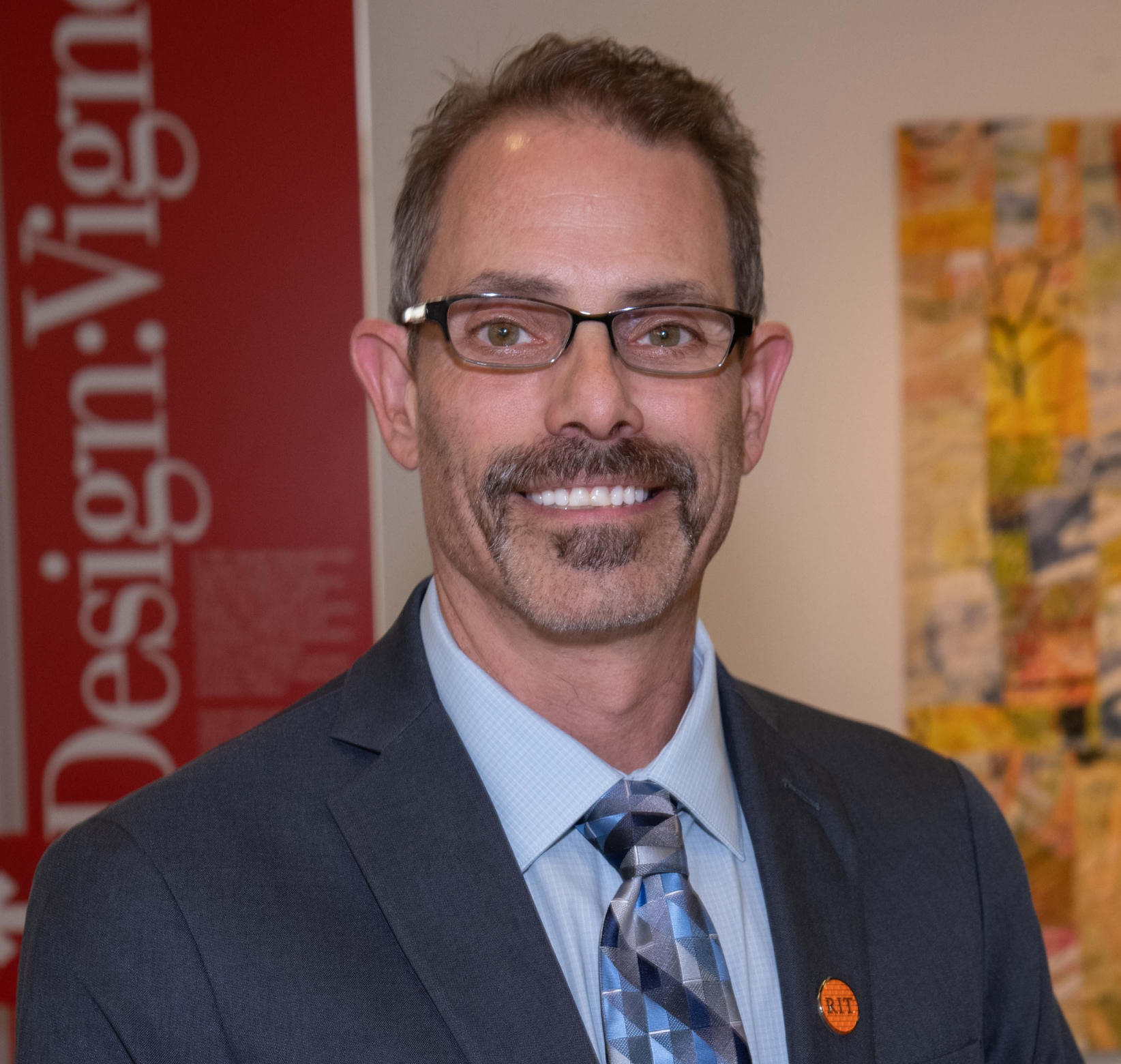}}]{Emmett J. Ientilucci}
Emmett J. Ientilucci (M'05-SM'17) received the B.S., M.S., and Ph.D. degrees in Imaging Science from the Rochester Institute of Technology, Rochester, NY, in 1996, 1999, and 2005, respectively.

Dr. Emmett Ientilucci is the Gerald W. Harris Endowed Professor in the Chester F. Carlson Center for Imaging Science, where he works in the Digital Imaging and Remote Sensing Laboratory. He has degrees in Optics and Imaging Science. He is the recipient of the 2020-21 Richard and Virginia Eisenhart Provost’s Award for Excellence in Teaching at RIT and is currently the IEEE Region 1 (NE USA) Area Chair and member of the International Honor Society IEEE-Eta Kappa Nu.

Dr. Ientilucci has been active in the field of remote sensing since 2000 and specifically in the area of spectral image analysis since 2004.  Prior to his university faculty position (in which he has taught courses in spectral image analysis, radiometry, remote sensing, geometrical optics, photo science, and metrology), he was a postdoctoral research fellow for the Intelligence Community.

His past and present research activities are in general remote sensing, spectral image processing and exploitation, hyperspectral target detection, shadow detection and mitigation, radiative transfer, radiometric hardware and calibration, atmospheric compensation, and landmine detection.

Dr. Ientilucci has 117 publications in the field of remote sensing.  He has served as a referee on 18 scientific journals, including being an Associate Editor (AE) for Optical Engineering and a current AE for GSRL. He has been a program reviewer for NASA, the Department of Defense (DOD), and is Chair for both the SPIE (Society for Optics and Photonics) Imaging Spectrometry Conference in San Diego, CA., and the Western NY Geoscience and Remote Sensing Society (GRSS). From 2016-2026 he has been the chair/co-chair of the IEEE GRSS UAV STRATUS Conference. He is a member of the International Society of Explosives Engineers (ISEE), Optica, and Senior member of both IEEE and SPIE.
\end{IEEEbiography}

\clearpage
\onecolumn
\normalsize
\setcounter{section}{0}
\setcounter{subsection}{0}
\setcounter{table}{0}
\setcounter{figure}{0}
\renewcommand{\thesection}{S\arabic{section}}
\renewcommand{\thesubsection}{S\arabic{section}.\arabic{subsection}}
\renewcommand{\thetable}{S\arabic{table}}
\renewcommand{\thefigure}{S\arabic{figure}}
\renewcommand{\theHsection}{supp.\arabic{section}}
\renewcommand{\theHsubsection}{supp.\arabic{section}.\arabic{subsection}}
\renewcommand{\theHtable}{supp.\arabic{table}}
\renewcommand{\theHfigure}{supp.\arabic{figure}}

\begin{center}
{\LARGE\bfseries Supplementary Material\par}
\vspace{0.5em}
{\large\bfseries SULAND\_v2: A Refined RGB Dataset and Deep Learning Object Detection Benchmark for UAV/UGV-Based SUrface LANDmine Detection Under Domain Shift\par}
\vspace{0.75em}
{\normalsize Sagar Lekhak, Prasanna Reddy Pulakurthi, Lalit Joshi, Ramesh Bhatta, and Emmett J. Ientilucci\par}
\end{center}
\vspace{0.8em}
This supplementary material provides additional information supporting the
SULAND\_v2 dataset audit and detector benchmark. It includes
(i)~representative folder- and sample-level annotation issues identified in
SULAND\_v1, together with a summary of the affected folders for each error
category, and (ii)~training-convergence results used to assess the adequacy of
the predefined training schedules for the evaluated detector configurations.
 A detailed sample-level audit spreadsheet, containing the affected folders, sample identifiers, error categories, target classes, and correction records, is released with the SULAND\_v2 dataset at \url{https://huggingface.co/datasets/SagarLekhak/SULAND_v2_RGB_Surface_Landmine_Dataset}. This material
complements the main manuscript without repeating its primary methodology or
benchmark results.

\section{Folder- and Sample-Level Audit Records}
\label{suppsec:folder_sample_audit}

This section provides a folder- and sample-level summary of
annotation issues identified in the original SULAND dataset, denoted
SULAND\_v1 in the main manuscript. The seven error categories defined in the
main paper are missing or incomplete annotations, false positive annotations,
mislocalized bounding boxes, inconsistent partial-visibility criteria,
non-representative artifacts, image-quality degradation, and class-ID
mismatch.

Table~\ref{tab:supp_folder_sample_audit} focuses on the five categories for
which representative affected folders and sample ranges can be listed
directly. Mislocalized bounding boxes and the OOD class-ID mismatch are not
included in the table because they affected a substantially larger portion of
the dataset. During the construction of SULAND\_v2, all samples were manually
reannotated, with bounding boxes tightened or otherwise corrected where
necessary. Localization changes can therefore be examined by comparing the
corresponding annotations in SULAND\_v1 and SULAND\_v2. Similarly, the
class-ID convention was corrected throughout the OOD subset and is described
separately in the main manuscript. The cases listed below are representative
rather than exhaustive.

\footnotesize
\setlength{\LTpre}{0.4em}
\setlength{\LTpost}{0.4em}
\renewcommand{\arraystretch}{1.12}
\setlength{\tabcolsep}{4pt}

\begin{longtable}{
    P{0.21\textwidth}
    P{0.23\textwidth}
    P{0.13\textwidth}
    P{0.35\textwidth}
}
\caption{Representative folder- and sample-level issues identified in
SULAND\_v1.}
\label{tab:supp_folder_sample_audit}\\

\toprule
\textbf{Error Category} &
\textbf{Folder/Samples} &
\textbf{Target} &
\textbf{Observed Issue} \\
\midrule
\endfirsthead

\multicolumn{4}{c}{\tablename~\thetable\ (continued)}\\
\toprule
\textbf{Error Category} &
\textbf{Folder/Samples} &
\textbf{Target} &
\textbf{Observed Issue} \\
\midrule
\endhead

\midrule
\multicolumn{4}{r}{Continued on next page}\\
\endfoot

\bottomrule
\endlastfoot

Missing or incomplete annotations &
\texttt{ITA-v1-0} &
Visible target &
The target is present, but the corresponding label file is empty. \\

Missing or incomplete annotations &
\texttt{ITA-v1-44--56} &
PMA-2 &
Visible PMA-2 targets are omitted from the released annotations. \\

Missing or incomplete annotations &
\texttt{ITA-v4-258--294} &
Both classes &
Both targets are visible in several frames, but only one is annotated in part
of the sequence. \\

Missing or incomplete annotations &
\texttt{ITA-v17-85--95}; comparison with \texttt{ITA-v17-99} &
PMA-2 &
The target is visible before annotation begins; sample 95 is unannotated,
whereas a visually comparable target is annotated in sample 99. \\

Missing or incomplete annotations &
\texttt{ITA-v21-0--8} &
PFM-1 &
The target is visible before the released annotations begin. \\

Missing or incomplete annotations &
\texttt{ITA-v21-770} &
Visible target &
The target is visible, but the corresponding \texttt{.txt} label file is
entirely missing. \\

Missing or incomplete annotations &
\texttt{ITA-v24-615--621} &
PFM-1 &
The target is visible before annotation begins at sample 622. \\

Missing or incomplete annotations &
\texttt{ITA-v25-317}; \texttt{ITA-v31-0} &
Visible target &
Clearly visible targets are omitted in isolated samples. \\

Missing or incomplete annotations &
\texttt{ITA-v32-35--46}; \texttt{ITA-v32-116--119} &
PMA-2; PFM-1 &
Visible PMA-2 and PFM-1 targets, respectively, are omitted. \\

Missing or incomplete annotations &
\texttt{ITA-v34-0--10}; \texttt{ITA-v34-39--45} &
Visible target &
Visible targets are omitted in two sample ranges. \\

\addlinespace
False positive annotations &
\texttt{ITA-v1-43} &
None &
An annotation is present although no valid target is visible. \\

False positive annotations &
\texttt{ITA-v14-86--146}, including \texttt{ITA-v14-100} &
None &
Bounding boxes persist across frames after the target is no longer visible. \\

False positive annotations &
\texttt{ITA-v37-591--592} &
None &
A PMA-2 annotation is present although no clear target is visible. \\

\addlinespace
Inconsistent partial-visibility criteria &
\texttt{ITA-v18-235} vs.\ \texttt{ITA-v16-54--61} &
PFM-1 &
A partially visible target is annotated in \texttt{ITA-v18-235}, whereas
comparable appearances are omitted in \texttt{ITA-v16}. \\

Inconsistent partial-visibility criteria &
\texttt{ITA-v20-63} vs.\ \texttt{ITA-v20-498} &
Visible target &
Comparable partial target appearances are treated differently within the
same folder. \\

Inconsistent partial-visibility criteria &
\texttt{ITA-v17-522--523} vs.\ \texttt{ITA-v17-329--330} &
PFM-1 &
A small visible target portion is annotated in one pair and omitted in the
comparable pair. \\

Inconsistent partial-visibility criteria &
\texttt{ITA-v22}, \texttt{ITA-v23}, \texttt{ITA-v29},
\texttt{ITA-v31}, and \texttt{ITA-v35} &
PMA-2 &
Highly occluded or weakly discernible targets are treated inconsistently
across folders. \\

\addlinespace
Non-representative artifacts &
\texttt{ITA-v14-32} &
PMA-2 &
A white marker is present near the target. \\

Non-representative artifacts &
\texttt{ITA-v32-20} &
PFM-1 &
A white marker or sticker is visible near the target. \\

Non-representative artifacts &
\texttt{ITA-v32-120--198}; \texttt{ITA-v32-231} &
Visible targets &
Marker-related artifacts occur repeatedly within the folder. \\

Non-representative artifacts &
\texttt{ITA-v1}, \texttt{ITA-v4}, \texttt{ITA-v6},
\texttt{ITA-v9}, \texttt{ITA-v11}, \texttt{ITA-v12}, and
\texttt{ITA-v13} &
PMA-2 &
Artificial markers or flags are attached to or positioned near PMA-2 targets. \\

\addlinespace
Image-quality degradation &
\texttt{ITA-v30-706--717}, including \texttt{ITA-v30-707} &
PFM-1 &
Severe motion blur reduces target visibility and obscures its boundaries. \\

\end{longtable}

\normalsize

\section{Affected Folders by Error Category}
\label{suppsec:affected_folders}

\subsection{Missing or Incomplete Annotations}
\label{suppsubsec:missing_incomplete}

\noindent\textbf{Affected folders:}
\texttt{ITA-v1}, \texttt{ITA-v4}, \texttt{ITA-v17}, \texttt{ITA-v21},
\texttt{ITA-v24}, \texttt{ITA-v25}, \texttt{ITA-v31}, \texttt{ITA-v32},
and \texttt{ITA-v34}.

\subsection{False Positive Annotations}
\label{suppsubsec:false_positive}

\noindent\textbf{Affected folders:}
\texttt{ITA-v1}, \texttt{ITA-v14}, and \texttt{ITA-v37}.

\subsection{Mislocalized Bounding Boxes}
\label{suppsubsec:mislocalized_boxes}

\noindent\textbf{Affected scope:}
All released IID and OOD folders were reviewed during reannotation, and
bounding boxes were redrawn or tightened where required. A representative
case from SULAND\_v1 is \texttt{ITA-v4-329}.

\subsection{Inconsistent Partial-Visibility Criteria}
\label{suppsubsec:partial_visibility}

\noindent\textbf{Affected folders:}
\texttt{ITA-v16}, \texttt{ITA-v17}, \texttt{ITA-v18}, \texttt{ITA-v20},
\texttt{ITA-v22}, \texttt{ITA-v23}, \texttt{ITA-v24}, \texttt{ITA-v29},
\texttt{ITA-v31}, and \texttt{ITA-v35}.

\subsection{Non-Representative Artifacts}
\label{suppsubsec:artifacts}

\noindent\textbf{Affected folders:}
\texttt{ITA-v1}, \texttt{ITA-v4}, \texttt{ITA-v6}, \texttt{ITA-v9},
\texttt{ITA-v11}, \texttt{ITA-v12}, \texttt{ITA-v13}, \texttt{ITA-v14},
and \texttt{ITA-v32}.

\subsection{Image-Quality Degradation}
\label{suppsubsec:image_quality}

\noindent\textbf{Affected folder:}
\texttt{ITA-v30}, particularly samples \texttt{ITA-v30-706--717}.

\subsection{Class-ID Mismatch}
\label{suppsubsec:class_id_mismatch}

\noindent\textbf{Affected folders:}
OOD folders \texttt{US1--US10}. The released OOD labels use the reverse of
the IID class convention, and the class IDs were corrected across the OOD
subset in SULAND\_v2.

\section{Training Convergence and Schedule Adequacy}
\label{suppsec:training_convergence}

\begin{figure}[!t]
  \centering
  \includegraphics[width=\textwidth]{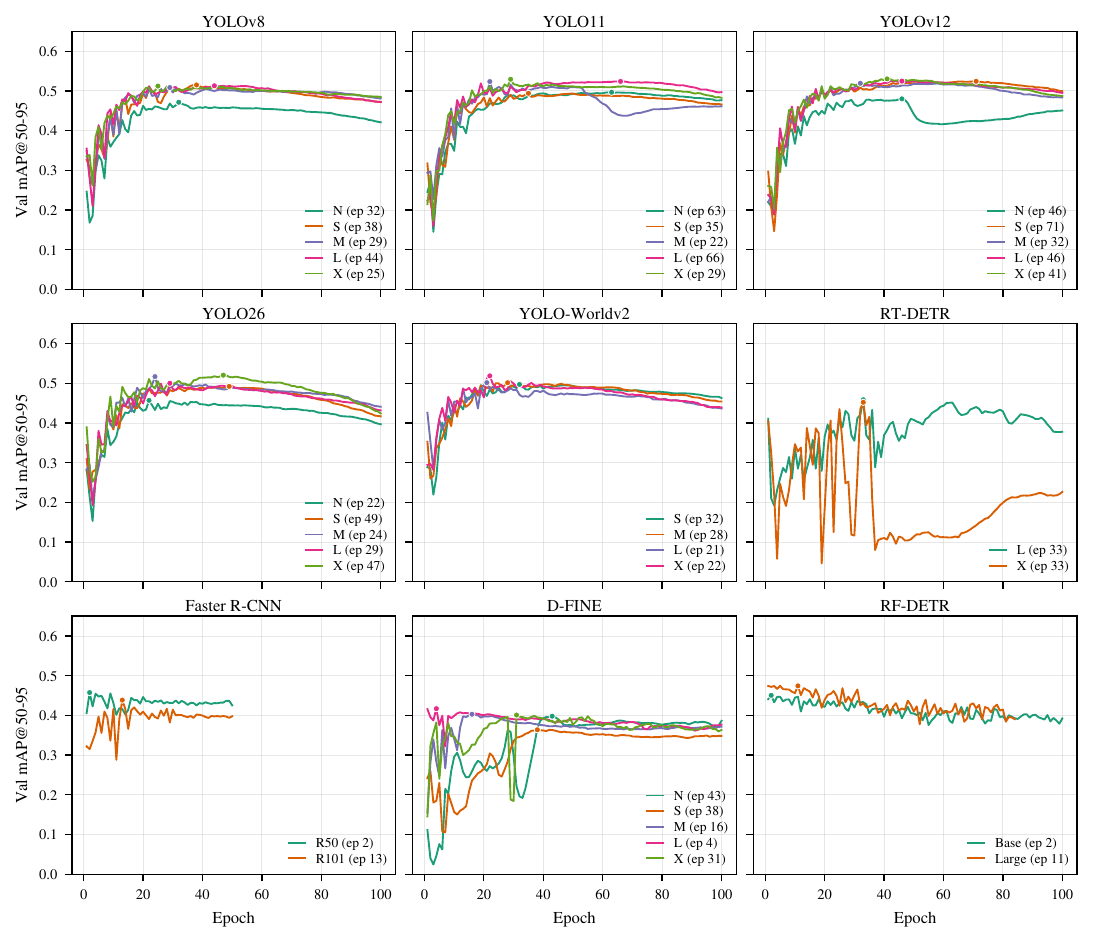}
  \caption{Validation mAP@50:95 over the training epochs for all benchmarked
  configurations, grouped by detector family. The marker identifies the epoch
  with the highest validation mAP@50:95, whose checkpoint was retained for
  final evaluation. The scheduled training length was 100 epochs for all
  configurations except the larger Faster R-CNN variants, which used
  50 epochs.}
  \label{fig:training_convergence}
\end{figure}

This section evaluates whether the predefined training schedules were
sufficient for the benchmarked detector configurations. It is included to
verify that the reported results were not affected by models being evaluated
before their validation performance had stabilized, particularly for
transformer-based detectors.

All configurations were trained for 100 epochs, except the larger Faster
R-CNN variants, which were trained for 50 epochs. For each configuration, the
checkpoint with the highest validation mAP@50:95 was retained for final
evaluation. Figure~\ref{fig:training_convergence} shows the corresponding
validation trajectories over the complete training schedules, grouped by
detector family.

Across the evaluated configurations, validation performance reaches a maximum
or stable plateau before training ends, including for RT-DETR, D-FINE, and
RF-DETR. These results indicate that the adopted schedules were adequate for
the unified comparison protocol. Detector-specific tuning or longer schedules
may further improve individual models, but such optimization was outside the
scope of this benchmark.

\end{document}